\documentclass[10pt,twocolumn,letterpaper]{article}

\usepackage[pagenumbers]{cvpr} 

\usepackage{times}
\usepackage{epsfig}
\usepackage{graphicx}
\usepackage{amsmath}
\usepackage{amssymb}
\usepackage{xcolor}         
\usepackage{float}
\usepackage{caption}
\usepackage{amsfonts,amssymb}
\usepackage{mathrsfs}
\usepackage{tabularx}
\usepackage{wrapfig}
\usepackage{verbatim}
\usepackage{dsfont}
\usepackage{tablefootnote}
\usepackage[stable]{footmisc}
\usepackage{booktabs}
\usepackage{multicol}
\usepackage{multirow}
\usepackage{color}
\usepackage{subcaption}
\usepackage{pythonhighlight}
\usepackage{algpseudocode}
\usepackage{algorithm}
\usepackage{physics}

\definecolor{todocolor}{RGB}{255,0,00}

\definecolor{jiapeng}{rgb}{0.2, 0.4,0.9}
\newcommand{\jiapeng}[1]{\textcolor{jiapeng}{\emph{jpt:~{#1}}}}

\definecolor{kai}{rgb}{0.9, 0.1, 0.9}

\definecolor{phg}{rgb}{0.7, 0.7, 0.2}

\definecolor{cyin}{rgb}{1.0, 0.0, 0.0}

\renewcommand{\paragraph}[1]{\smallskip\noindent\textbf{#1}}

\definecolor{sim}{RGB}{128,0,128}
\definecolor{interp}{RGB}{255, 169, 0}
\definecolor{mytbcol}{RGB}{175,227,246}

\definecolor{tabfirst}{rgb}{0.625, 0.875, 0.6} 
\definecolor{tabsecond}{rgb}{0.90, 0.975, 0.875} 
\definecolor{tabthird}{rgb}{1, 1, 0.7} 

\newcommand{\tabfirst}[1]{\colorbox{tabfirst}  {#1}}
\newcommand{\tabsecond}[1]{\colorbox{tabsecond} {#1}}

\definecolor{cvprblue}{rgb}{0.21,0.49,0.74}
\usepackage[pagebackref,breaklinks,colorlinks,allcolors=cvprblue]{hyperref}

\usepackage{pifont}
\newcommand{\cmark}{\ding{51}}%
\newcommand{\xmark}{\ding{55}}%
\usepackage{tabularx}

\def\paperID{8753} 
\def\confName{CVPR}
\def\confYear{2026}

\title{FactorPortrait: Controllable Portrait Animation via Disentangled \\ Expression, Pose, and Viewpoint}

\author{
Jiapeng Tang$^{1,2}$ \quad Kai Li$^1$ \quad Chengxiang Yin$^1$ \quad Liuhao Ge$^1$ \quad Fei Jiang$^1$ \quad Jiu Xu$^1$ \\
\quad Matthias Nie{\ss}ner$^2$ \quad Christian H{\"a}ne$^1$ \quad Timur Bagautdinov$^1$ \quad Egor Zakharov$^1$ \quad Peihong Guo$^1$ \\
 $^{1}$ Meta Reality Labs
  $^{2}$ Technical University of Munich
}

\begin{document}
\twocolumn[{%
	\renewcommand\twocolumn[1][]{#1}%
	\maketitle
	\begin{center}
            \vspace{-8mm}
		\centerline{
                \includegraphics[width=\linewidth]{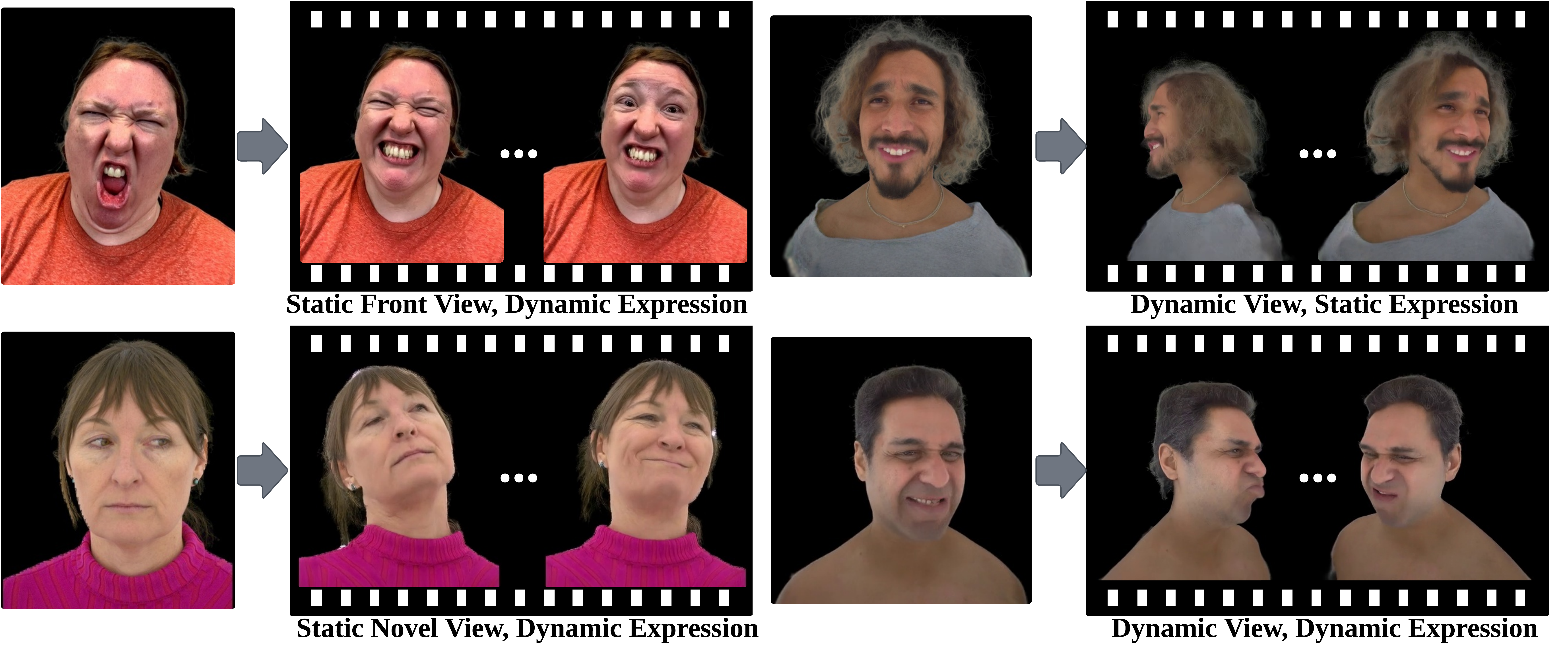}
		}
        \vspace{-2mm}
        \captionof{figure}{Given a single portrait image, FactorPortrait generates vivid portrait animations featuring complex facial dynamics, and precise, flexible camera control. Our method supports a wide range of controllable combinations, including viewpoint, pose, and expression.
        } 
        \label{fig:teaser}
	\end{center}
    \vspace{-3mm}
}]


\begin{abstract}
We introduce FactorPortrait, a video diffusion method for controllable portrait animation that enables lifelike synthesis from disentangled control signals of facial expressions, head movement, and camera viewpoints. Given a single portrait image, a driving video, and camera trajectories, our method animates the portrait by transferring facial expressions and head movements from the driving video while simultaneously enabling novel view synthesis from arbitrary viewpoints.
We utilize a pre-trained image encoder to extract facial expression latents from the driving video as control signals for animation generation. Such latents implicitly capture nuanced facial expression dynamics with identity and pose information disentangled, and they are efficiently injected into the video diffusion transformer through our proposed expression controller.
For camera and head pose control, we employ Pl{\"u}cker ray maps and normal maps rendered from 3D body mesh tracking.
To train our model, we curate a large-scale synthetic dataset containing diverse combinations of camera viewpoints, head poses, and facial expression dynamics. Extensive experiments demonstrate that our method outperforms existing approaches in realism, expressiveness, control accuracy, and view consistency. \href{https://tangjiapeng.github.io/FactorPortrait}{Project Page}
\end{abstract}    
\section{Introduction}
\label{sec:intro}
Generating lifelike portrait animation from a single image has wide applications in virtual and augmented reality, film, education, and entertainment. However, it is an inherently ambiguous problem due to the limited information present in a single image. High-fidelity appearances and realistic facial motions generation without identity shift are key challenges.

Generative Adversarial Networks (GANs)~\cite{goodfellow2020generative} have shown promise in generating such animations. GAN-based methods~\cite{nirkin2019fsgan, siarohin2019first, cheng2022videoretalking, doukas2021headgan, drobyshev2022megaportraits, zhang2023metaportrait} can generate richer facial details than conventional video animation methods~\cite{alexander2010digital, kirschstein2023nersemble, li2017learning, lombardi2021mixture}, but exhibit poor generalization to unseen identities, have visual artifacts, motion distortion, and lack of sufficient control over facial expressions.
Pre-trained foundational diffusion priors, \eg Stable Diffusion~\cite{rombach2022high} and Wan~\cite{wan2025}
have shown promising results when adapted to facial animation generation~\cite{guo2023animatediff, guo2024real, xie2024x, yang2024megactor}. 
2D facial landmarks for representing facial expressions \cite{ma2024followyouremoji,  wei2024aniportrait} can only capture coarse movements of facial features, and 3D Morphable Model (3DMM) as dense geometry guidance~\cite{chen2024morphable, tang2025gaf, prinzler2025joker, Taubner2024CAP4DCA} is unable to capture fine-grained details such as wrinkles.
Moreover, existing methods are restricted to frontal viewpoints and lack continuous viewpoint control, limiting their applicability in VR/AR applications. 
To this end, we propose FactorPortrait, a video diffusion method that enables life-like portrait video synthesis from disentangled control signals of facial expression, head movement and camera viewpoints.
We designed an expression controller that efficiently injects expression information into the DiT~\cite{peebles2023scalable} based video diffusion network with minimal learnable parameters.
For pose control, we utilize parametric body mesh tracking to obtain body meshes that are rendered into normal maps as spatially-aligned dense conditioning input.
We adopt Pl{\"u}cker ray maps to represent viewpoint. Finally, we fuse identity appearance cues, template mesh normal maps, and ray maps in a condition fusion layer before the DiT network.

Controllability in video diffusion models requires a supervised fine-tuning dataset with accurate annotations for each control factor in monocular videos, covering diverse combinations of viewpoint, pose, and expression. 
In-the-wild portrait videos could provide the needed variety on head poses and expressions, but they are often captured from fixed, frontal viewpoints. For the videos with camera movements, accurately recovering head articulation and camera motions from in-the-wild monocular videos remains challenging due to local minima in rigid and non-rigid tracking. 
One might consider rendering continuous views from static 3D head assets to augment continuous camera motions in the training dataset. However, this approach only generates videos with static expressions or poses.
%
To address this limitation, we employ a synthetic dataset containing video renderings from high-quality Gausssians-based animatable head avatars reconstructed from dense multi-view studio captures~\cite{martinez2024codec}. These avatars enable novel view renderings along arbitrary continuous camera trajectories while simultaneously allowing expression and pose changes during camera movements, simulating realistic observation scenarios in VR/AR applications.

As shown in Fig.~\ref{fig:teaser}, our method generates high-quality portrait animations with accurate identity preservation and complex facial expressions, while enabling precise, flexible camera control: (1) static frontal viewpoint with dynamic pose and expression; (2) static novel viewpoints with dynamic pose and expression; (3) static pose and expression with dynamic viewpoint; and (4) simultaneous dynamic pose, expression, and viewpoint.

The contributions of this paper can be summarized as:
\begin{itemize}
\item We introduce a controllable portrait video diffusion model that enables flexible combinations of facial expressions, camera viewpoints, and head poses.
\item We propose a data curation strategy that augments monocular videos with synthetic renderings, enabling continuous view synthesis with both static and dynamic expressions and poses.
\item We design an expression controller network that efficiently injects latent expression codes into DiT with minimal learnable parameters while capturing complex facial expressions.
\end{itemize}
Extensive experiments demonstrate that our method outperforms state-of-the-art portrait animation methods across different datasets and control modes.
%

\section{Related Work}
\label{sec:related}
%

\subsection{Portrait Video Animation}
%
%
%
%
Portrait video animation methods can be divided into non-diffusion and diffusion-based approaches.

%
%
Non-diffusion work~\cite{siarohin2019fomm, wang2021facevid2vid, guo2024liveportrait} utilized implicit keypoints as motion representations to warp source portraits, while others~\cite{lin2022ganinversion, xie2024x} encoded expressions as latent vectors and injected them into generator networks for feature-space manipulation.
%
%
To incorporate geometric priors, some methods integrated 3DMM~\cite{blanz1999morphable} into GANs~\cite{ververas2022f3agan, behrouzi2024maskrenderer} or employed 3DMM blendshapes as motion representations~\cite{ma2024gaussian, chen2024bringyourown}.
%
%
However, these non-diffusion-based approaches lack robust priors for extreme poses and expressions, and warping-based strategies fail to achieve 3D consistency and high rendering quality under large head and body movements.

%

%
Recent diffusion-based approaches~\cite{ma2024followyouremoji, xu2025hunyuanportrait, gao2025learn2control} achieved significant progress in portrait animation by adapting visual foundation models~\cite{rombach2022high, baldridge2024imagen, blattmann2023align, kong2024hunyuanvideo}. 
FADM~\cite{Zeng2023FaceAW} pioneered diffusion-based portrait animation, followed by methods~\cite{karras2023dreampose, wei2024aniportrait, xu2024magicanimate, ma2024followyouremoji} that fine-tune Stable Diffusion~\cite{rombach2022high} for human portrait animation. 
%
%
To achieve temporal consistency, subsequent methods~\cite{blattmann2023stable, wang2024vividpose, guo2024animatediff, hu2024animateanyone, zhu2024champ, wang2024disco, tong2024musepose, tian2024emo, wang2024unianimate} leveraged image or video diffusion models in an end-to-end fashion for temporally coherent portrait video generation. They mitigated background jitter issues while enabling superior identity generalization. 
%
%
%
DiffusionRig~\cite{ding2023diffusionrig}, DiffusionAvatars~\cite{Kirschstein2023DiffusionAvatarsDD}, ConsistentAvatar~\cite{Yang2024ConsistentAvatarLT}, and Stable Video Portraits~\cite{Ostrek2024StableVP} generated avatar animations but required subject-specific training.
For expression control, some methods~\cite{ma2024followyouremoji, wei2024aniportrait} used 2D facial landmarks, which provide only coarse and inaccurate control. Others~\cite{chen2024morphable, tang2025gaf, prinzler2025joker, Taubner2024CAP4DCA} leveraged 3DMM reconstruction to guide image or video diffusion models. However, the limited representation capacity of PCA-based parametric models and fixed template meshes makes it difficult to capture nuanced facial expressions, such as wrinkles. HunyuanPortrait~\cite{xu2025hunyuanportrait} recently introduced implicit expression latents for UNet-based video diffusion models~\cite{blattmann2023stable} using additional cross-attention layers, but at a high computational cost.
In contrast, we present an expression controller that injects expression latents into DiT~\cite{peebles2023scalable} via Adaptive Layer Normalization layers~\cite{ba2016layer, perez2018film}, introducing only minimal learnable parameters.

%
%
%

\subsection{Camera Conditioned Diffusion Models}
%
Early approaches~\cite{Liu2023Zero1to3ZO, Shi2023Zero123AS} introduced camera pose conditioning into pretrained text-to-image diffusion models for novel view synthesis.
To improve multi-view consistency, later methods~\cite{liu2023syncdreamer, Shi2023MVDreamMD, Wang2023ImageDreamIM, Gao2024CAT3DCA, Taubner2024CAP4DCA} employed 3D-aware attention mechanisms to jointly denoise multiple views, thereby enforcing consistency across them.
However, these image-based models lack temporal priors, which leads to inconsistency when generating views with significant viewpoint changes.
%
%
Recent video-based work~\cite{Wu2024CAT4DCA, Zhou2025StableVC, Bai2025ReCamMasterCG, Mark2025TrajectoryCrafterRC, Yu2024ViewCrafterTV, Ren2025Gen3C3W} finetuned video diffusion models for camera control along continuous trajectories, achieving smoother view transitions and improved temporal consistency. MVPerformer~\cite{zhi2025mv} jointly denoised multi-view human videos and enabled novel view rendering through 4D reconstruction from monocular video.
However, these methods focus solely on camera control and do not generate novel facial expressions or head movement beyond the input. Many approaches~\cite{Bai2025ReCamMasterCG, Mark2025TrajectoryCrafterRC, zhi2025mv} require a monocular video as input.
In contrast, our approach accepts a single image as input and enables comprehensive portrait animation with precise control over various combinations of dynamic camera viewpoints, facial expressions, and body poses.

\section{Dataset Curation}
\label{sec:dataset}

\begin{figure*}
    \vspace{-3mm}
    \centering
    \includegraphics[width=\linewidth]{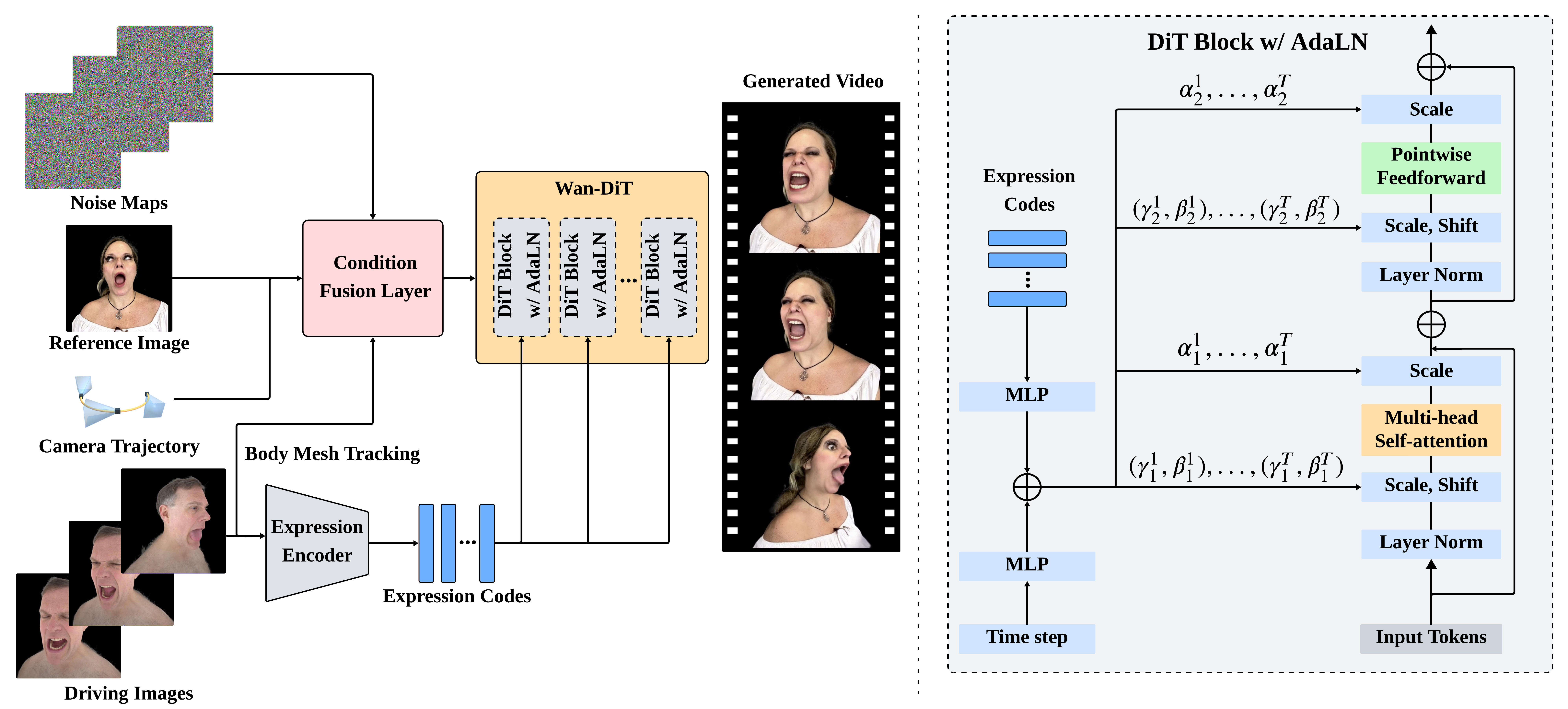}
    \vspace{-7mm}
    \caption{\textbf{Pipeline Overview.} Our method generates a video of the reference subject animated by the body pose and facial expressions from the driving images, while following the specified camera trajectory. The model consists of three main components: (1) a condition fusion layer that combines noise maps, the reference image, and camera pose annotations, and body mesh tracking as input to DiT; (2) an expression encoder that extracts and aggregates per-frame expression codes from the driving images; and (3) a video diffusion model based on Wan-DiT blocks with adaptive layer normalization (AdaLN), which applies scale and shift transformations conditioned on the per-frame expression codes and frame-agnostic timestep embedding.
    } 
    \label{fig:overview}
    \vspace{-3mm}
\end{figure*}

 To achieve fully disentangled control over multiple signals, the ideal approach is to acquire large-scale videos that encompass all possible dynamics simultaneously. However, collecting such comprehensive data at scale is often impractical, primarily in data storage and computational resources.
One possible approach is to recover all the disentangled dynamics from monocular videos, including head poses, facial expressions, and camera parameters. However, accurately solving rigid and non-rigid tracking, remains a significant challenge. 
In this section, we present the carefully designed dataset curation strategies that deliver disentangled and accurate dynamics for model training, where each dataset covers a subset of control signals at a time.

\subsection{Real Data} 
Due to the aforementioned data constraints, direct joint training for video synthesis with fully disentangled controls is intractable. To address this, we leverage both monocular iPhone captures and multi-view studio recordings to incrementally develop these capabilities.

\noindent \textbf{Phone Capture.} We utilize a monocular iPhone video dataset comprising 11,976 identities, with on-average 4,000 frames per capture from 30 videos, at a resolution of 1440x1080. The videos include a variety of actions such as head rotation, brief expressions, and speech. We primarily leverage the rich identities and diverse facial dynamics.

\noindent \textbf{Studio Capture.} The multi-view studio dataset, similar to the ones in~\cite{cao2022authentic, li2024uravatar}, includes 1414 identities recorded with 78 synchronized 2K cameras, each providing approximately 4,000 frames across diverse facial expressions, head movements, and gaze directions. For each capture, 11 views are randomly sampled to balance coverage and computational efficiency. We retain 612 raw captures for training expression dynamics and novel view synthesis.

\subsection{Synthetic Data} 
%
We propose using animatable head avatars to generate synthetic videos with disentangled signals, by rendering two distinct types of videos: (1) \emph{ViewSweep}: contains static expressions and varying camera trajectories; (2) \emph{DynamicSweep}: contains simultaneous changes in both facial dynamics and camera viewspoints. This synthetic data explicitly disentangles camera motion from portrait dynamics, enabling clear, independent supervision of each signal. 

\noindent \textbf{Gaussian Avatar Fitting.} We fit animatable Gaussian avatars for 802 studio captures, with disentangled expression code, camera view, and pose, similar to the universal prior model in~\cite{li2024uravatar, liu2025lucas} but without hair-specific control nor lighting input. An expression encoder~\cite{seamless_interaction} is used to extract latent expression codes. A hypernetwork conditioned on identity information generates person-specific bias maps. The final guide mesh and Gaussian parameters are produced for image rendering. 

\noindent \textbf{Re-rendering.} With the fitted animatable Gaussian avatars, we render arbitrary videos using desired expression codes, body poses, and novel cameras. This disentangles the camera motion and facial dynamics in the rendered videos, allowing for independent supervision of each control signal during model training.
\begin{itemize}
    \item \emph{ViewSweep}. For each identity, we randomly select a facial expression and design a camera trajectory (e.g., spin or spiral) with varied distance and look-at points. This yields 128 unique 100-frame sequences at 1024x1024 resolution per identity.
    \item \emph{DynamicSweep}. Rather than keeping expressions static during camera motion, facial expressions and body poses are sampled from random segments of the original capture. Each identity generates 32 unique 128-frame trajectories at 1024x1024 resolution.
\end{itemize}

\section{Controllable Portrait Animation}
\label{sec:controllable}


Our pipeline generates a video of the reference subject, controlled by the camera views, body poses, and facial expressions from the driving image sequence, as illustrated in Fig.~\ref{fig:overview}. We adapt video foundation prior model Wan~\cite{wan2025} as the backbone for the face domain task through supervised training. The encoders responsible for extracting disentangled identity, pose, expression, and camera information are described in Sec.~\ref{sec:condition}. Sec.~\ref{sec:diffusion} shows how to integrate these control signals into the Diffusion Transformers~\cite{peebles2023scalable} denoising framework. Our training strategy for a high-fidelity and fully disentangled control is introduced in Sec.~\ref{sec:progressive}.

\subsection{Disentangled Conditions}
\label{sec:condition}

Given a single portrait image $\mathbf{I}$ as reference, a camera trajectory $\mathbf{C}$, and a driving video~$\mathbf{D}$ of length $T$ (another identity or same identity), the goal is to generate the high-fidelity portrait video with temporal coherence. The generated video should: 1) preserve the identity and appearance of the reference image $\mathbf{I}$; 2) follow the camera trajectory $\mathbf{C}$ to render novel viewpoints; 3) inherit the expression variations of the driving video $\mathbf{D}$. To achieve disentangled control, we first extract conditions on identity, pose, viewpoint, and expression from disjoint inputs.

\noindent \textbf{Identity Condition.}
For identity preservation, we extract the latent features $\mathbf{z_I}$ from reference image $\mathbf{I}$ using the Wan-VAE encoder $\mathcal{E}$~\cite{wan2025}. Unlike identity embeddings from face recognition models~\cite{schroff2015facenet}, which often lose fine-grained appearance information, this VAE latent retrains rich low-level facial details, including skin texture, facial structure, hair, and other identity-specific characteristics. They are seamlessly integrated into the diffusion process via attention mechanisms within the same latent space.

\noindent \textbf{Pose Condition.}
To extract body pose from the driving video $\mathbf{D}$, we estimate the parametric body meshes~$\mathcal{M_D}$ via a feed-forward 3D human mesh method based on~\cite{khirodkar2024sapiens}, and render body meshes $\mathcal{M_D}$ into normal maps $\mathbf{N_D}$. The normal maps provide a dense, pixel-aligned representation of 3D body movements and is seamlessly integrated into the video diffusion models for pose control. Unlike rotation matrix or Euler angles, normal maps maintain spatial correspondence within the image domain. Thanks to the human body priors~\cite{khirodkar2024sapiens}, the body mesh estimation is robust even in challenging scenarios such as occlusions, providing us reliable pose conditioning.

\noindent \textbf{Camera Condition.}
Similar to previous work~\cite{chen2025v3d, he2025cameractrl}, Pl{\"u}cker coordinates are used to represent camera viewpoints in a continuous manner. For each driving frame $\mathbf{D}_i$, we compute the relative camera pose $\boldsymbol{\pi}_i$ between the driving frame $\mathbf{D}_i$ and the reference image $\mathbf{I}$. Pl{\"u}cker ray maps $\mathbf{R}_i$ are then constructed to encode both the direction and position of the rays from the target camera viewpoint. 
%

\noindent \textbf{Expression Condition.}
We utilize a pre-trained expression encoder~\cite{seamless_interaction} to extract 128-d expression latent codes $\{\mathbf{E}_{\mathbf{D}_i}\}$ from the driving video $\mathbf{D}$. Traditional methods for facial expression representations typically rely on facial landmarks or parametric models like  FLAME~\cite{flame2017}. However, they have significant limitations in capturing nuanced facial expressions, including micro-expressions, fine wrinkles, mouth interior and tongue movements. In our pipeline, the expression encoder is designed to overcome these challenges by implicitly capturing fine-grained, complex, and non-linear expression dynamics, while excluding identity information by an aligner encoder and frame latent encoder.
%

\subsection{Controllable Video Diffusion}
\label{sec:diffusion}
We now detail how to input these condition signals into the Wan~\cite{wan2025}-based video diffusion transformer with two key modules: condition fusion layer and expression controller.

\noindent \textbf{Video Diffusion Transformer.} Given an input video $\mathbf{V}\in\mathbb{R}^{T \times H \times W \times 3}$ with $T$ frames, Wan~\cite{wan2025} uses a causal video autoencoder to encode $\mathbf{V}$ into a compact spatiotemporal latent representation $\mathbf{z}=\mathcal{E}(\mathbf{V})\in\mathbb{R}^{l \times h \times w \times c}$, where $l=(T+3)//4$, $h=H//8$, $w=W//8$, and $c=16$ denote the temporal, height, width, and channel dimensions, respectively. 
Wan leverages flow matching~\cite{lipman2022flow} to learn a continuous-time ordinary differential equation (ODE) that transforms Gaussian noise $\boldsymbol{\epsilon}$ into the video latent $\mathbf{z}$, conditioned on input signals. During training, video latent $\mathbf{z}$ is gradually perturbed to produce noisy versions $\mathbf{z}_t$, and a denoising transformer is trained to predict the velocity field needed to recover the original latent structure.


\noindent \textbf{Condition Fusion Layer.}
\begin{figure}
    \centering
    \includegraphics[width=\linewidth]{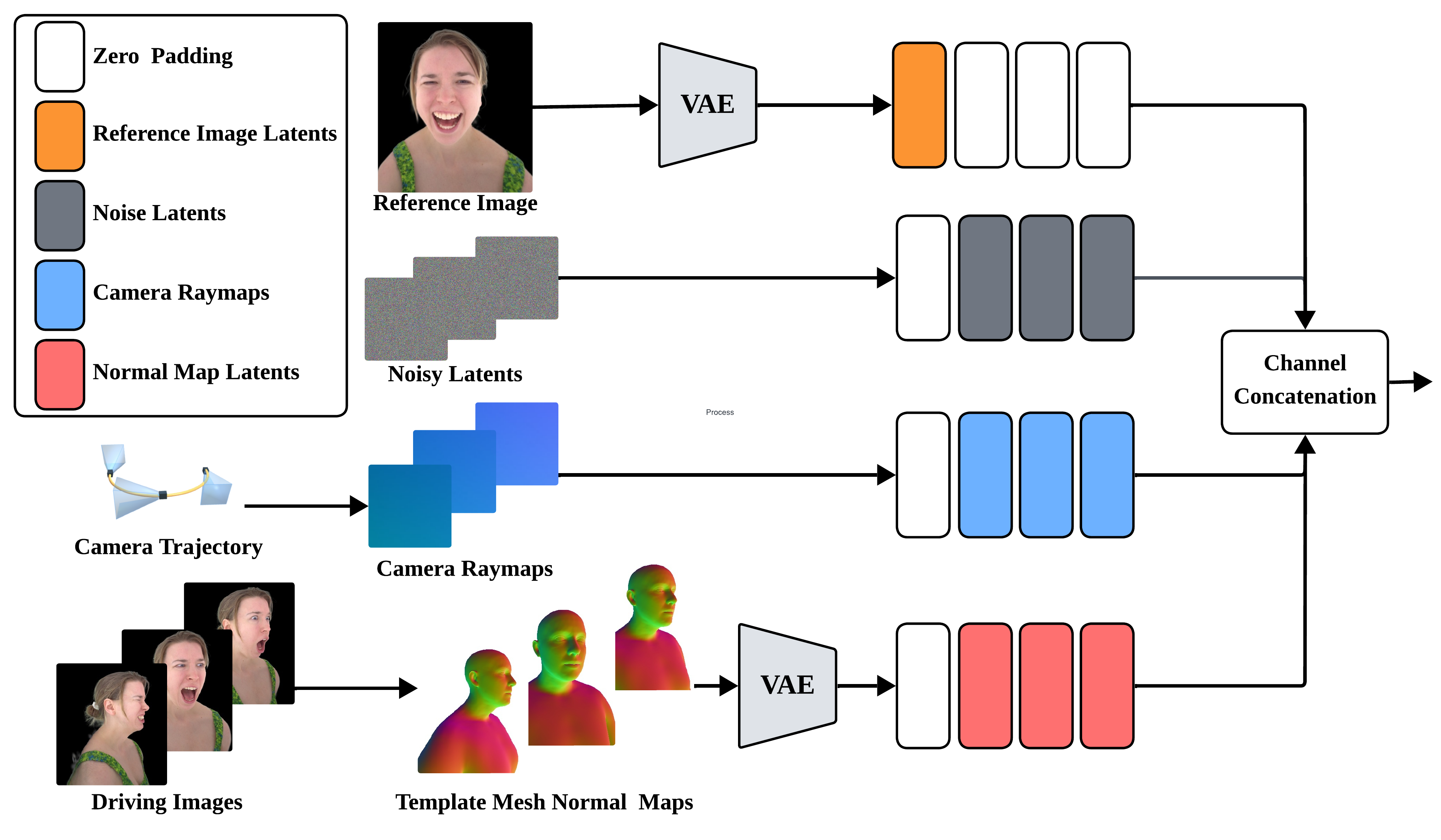}
    \vspace{-3mm}
    \caption{\textbf{Condition Fusion Layer.}
    We extract reference image latents, ray maps, and normal video latents from the template mesh to represent identity, viewpoint, and pose, respectively. They are concatenated with noise latents as input to the diffusion model.
    }
    \label{fig:inputlayer}
    \vspace{-3mm}
\end{figure}
As shown in Fig.~\ref{fig:inputlayer}, a unified input layer is introduced to fuse multiple conditioning signals via feature concatenation. Specifically, we combine noisy video latent~$\mathbf{z}_t$, reference image latents~$\mathbf{z_I}$, body normal maps latents~$\mathbf{N}$, and camera ray maps~$\mathbf{R}$ into a single feature.
\begin{itemize}
    \item Normal maps and ray maps are concatenated along the channel dimension to form a dense spatial signal.
    \item Reference image latent $\mathbf{z_I}$ is prepended to the noisy video latent $\mathbf{z}_t$ along the temporal dimension, treating it as the first frame of the sequence.
\end{itemize}

For reference frames, normal maps and ray maps are zero-padded and computed from the identity camera matrix, since there is no motion or viewpoint change. For generated frames, reference image features are set to zero, ensuring that each frame receives only its relevant conditioning signals. This asymmetric conditioning design helps the model to learn the relationship between the static reference and the dynamic generated sequence.

\begin{figure*}
    \centering
    \vspace{-3mm}
    \includegraphics[width=\linewidth,trim={0 5cm 0 2cm},clip]{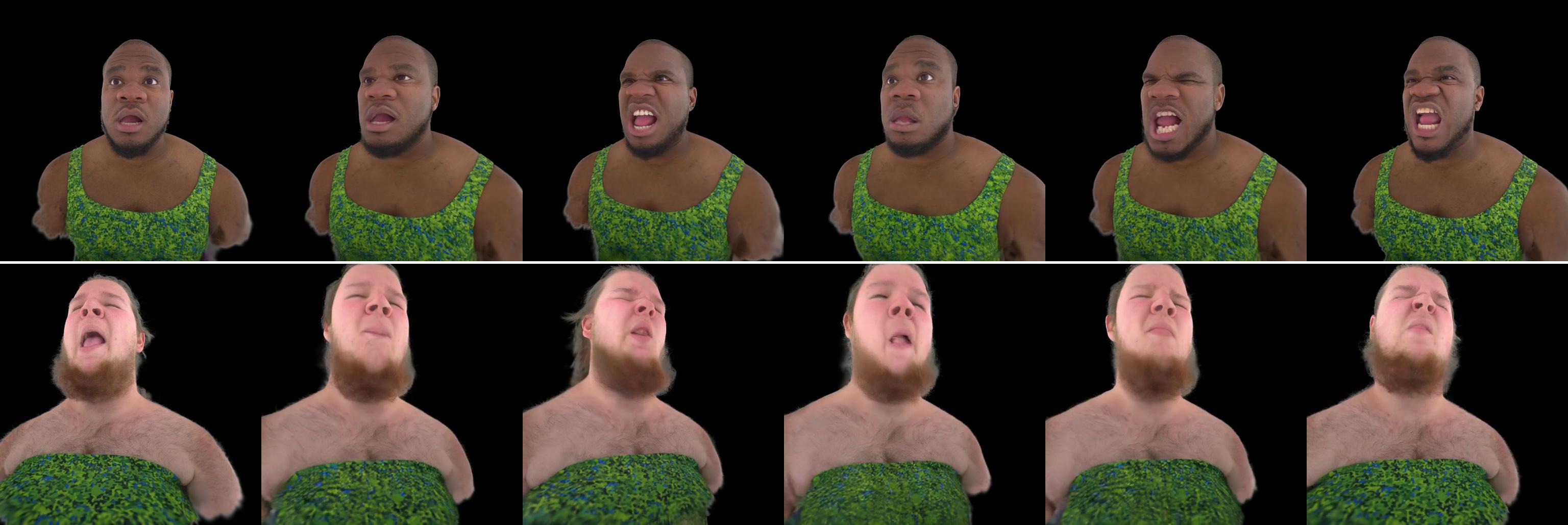}
    \begin{tabular}{p{80pt}p{80pt}p{70pt}p{80pt}p{70pt}p{90pt}}
        \centering Input & \hspace{8pt} C1 & C2 &  C3 & Full &  GT
    \end{tabular}
    \vspace{-5mm}
    \caption{\textbf{Qualitative Ablation Studies on the DynamicSweep dataset}. C1: without DynamicSweep during training, C2: without body normal maps as head pose conditions, C3: change expression latents to landmarks as expression conditions. 
    Each component is essential for achieving the desired view, head pose, and expression control in our method.}
    \label{fig:ablation}
    \vspace{-5mm}
\end{figure*}

\noindent \textbf{Expression Controller.}
To achieve precise expression control, a sequence of $T\times 128$ expression codes \{$\mathbf{E}_{\mathbf{D}_i}$\} are first processed using two self-attention layers, which aggregates temporal dependencies across frames and outputs features of size $T\times C$. They are grouped by chunks of 4 consecutive frames into features \{$\mathbf{e}_i$\} of size $(T+3)//4 \times 4C$, except for the first frame corresponding to the reference image, to align with the video latent for frame-wise conditioning.

We then add the compressed expression latent $\mathbf{e}_i$ to the shared timestep embedding $\mathbf{t}$, resulting in frame-specific timestep embeddings $\mathbf{t}_i=\mathbf{t}+\mathbf{e}_i$. These embeddings are used to predict shift and scale parameters that modulate the video latent at each frame via adaptive layer normalization (AdaLN)~\cite{perez2018film, ba2016layer, peebles2023scalable}.
\begin{equation}
\text{AdaLN}(\mathbf{z}_i, \mathbf{t}_i) = \gamma(\mathbf{t}_i) \cdot \frac{\mathbf{z}_i - \mu(\mathbf{z}_i)}{\sigma(\mathbf{z}_i)} + \beta(\mathbf{t}_i),
\end{equation}
where $\mathbf{z}_i$ denotes the latent features of the $i$-th frame, $\gamma(\cdot)$ and $\beta(\cdot)$ are learned functions that predict the scale and shift parameters from the frame-specific timestep embedding, $\alpha$ is the dimension-wise scaling parameters applied prior to any residual connections, and $\mu(\cdot)$ and $\sigma(\cdot)$ are the mean and standard deviation. This design allows the model to apply distinct expression conditions to each frame independently, while maintaining temporal coherence through the shared base timestep embedding. Compared to cross-attention layers, our design requires minimal computes with only a few additional learnable parameters.

\subsection{Progressive Training}
\label{sec:progressive}
\setlength{\tabcolsep}{3pt}
\begin{table}[tbp]
	\renewcommand\arraystretch{1.2}
	\begin{center}
		\begin{tabular}{*{5}{c}}
			\toprule
              & Stage 1 & Stage 2 & Stage 3 & Stage 4 \\
            \midrule
            PhoneCapture & 100\% & 60\% & 20\% & 20\% \\
            StudioCapture & - & 40\% & 20\% & 20\% \\
            ViewSweep & - & - & 30\% & 30\% \\
            DynamicSweep & - & - & 30\% & 30\% \\
            \midrule
            Timestamps & 13 & 25 & 49 & 81 \\




        \bottomrule
        \end{tabular}
        \caption{\textbf{Progressive training.} 
      	We gradually enhance the model’s ability to generate temporally smooth videos of increasing duration while supporting disentangled control of expression, pose, and viewpoint.
      } 
        \label{tab:train}
        \end{center}
        \vspace{-7mm}
\end{table}

As illustrated in Tab.~\ref{tab:train}, our video diffusion model is trained with a staged strategy, where each each stage is designed to address specific learning objectives.
\begin{itemize}
    \item \textbf{Stage 1} uses only the PhoneCapture data to focus on robust control over facial expressions and head pose, leveraging rich identities for identity preservation. 
    \item \textbf{Stage 2} incorporates the multi-view StudioCapture for novel view synthesis. This further bootstraps the learning on expression and pose, by explicit supervision of disentangled camera and facial dynamics.
    \item \textbf{Stage 3} adds the synthetic sequences, to disentangle camera motion by exposing the model to a wide variety of camera trajectories alone, or with simultaneous changes in expression and pose together.
    \item \textbf{Stage 4} maintains the same dataset ratio as Stage 3 but focuses on generating longer video sequences.
\end{itemize}

By gradually increasing the timestamps, our model starts from basic temporal transitions and incrementally learns more complex temporal changes. This curriculum-based approach helps the model build a strong foundation in short-term dynamics before tackling long-range ones.

\section{Experiment}
\label{sec:experiment}

\subsection{Implementation Details}
\label{SubSecImplementation}

\noindent \textbf{Datsets.}
We train our model using the combined dataset describe in Sec.~\ref{sec:dataset} and Tab.~\ref{tab:train}. For evaluation, we randomly sample 50 sequences of unseen identities from each dataset.

\noindent \textbf{Comparisons.}
We compare our method against recent state-of-the-art approaches for portrait animation. GAGAvatar~\cite{chu2024generalizable} reconstructs animatable Gaussian avatars from a single image and renders novel views and expressions. CAP4D~\cite{Taubner2024CAP4DCA} is a multi-image diffusion model guided by 3DMM tracking~\cite{flame2017}. HunyuanPortrait~\cite{xu2025hunyuanportrait} is a state-of-the-art video diffusion method for talking face generation.

\noindent \textbf{Evaluation Metrics.}
We evaluate the generated videos in four aspects: 1) \textit{Image quality}: PSNR, LPIPS, and SSIM measure pixel-level alignment and structural similarity, similar to~\cite{Gao2024CAT3DCA}; 2) \textit{Identity similarity}: CSIM computed from ArcFace embeddings~\cite{deng2018arcface} measures identity preservation; 3) \textit{Expression accuracy}: AED~\cite{siarohin2019first} measures the difference in 3DMM expression coefficients between generated and ground-truth images using Deep3DFaceRecon~\cite{deng2019accurate}; 4) \textit{Video quality}: FID~\cite{Liu2023Zero1to3ZO}, FVD~\cite{DBLP:journals/corr/abs-1812-01717}, and IQA~\cite{xu2025hunyuanportrait} assess temporal consistency and perceptual quality.

\noindent \textbf{Implementations.}
Our model is initialized from the pre-trained Wan 2.1-T2V models and trained on 64 GPUs with a batch size of 64. We employ a four-stage training strategy, where each stage is initialized from the previous stage. Stages 1 and 2 are trained for 20,000 iterations with a learning rate of $1 \times 10^{-4}$. Stage 3 uses a learning rate of $5 \times 10^{-5}$ for 20,000 iterations. Stage 4 uses a learning rate of $2 \times 10^{-5}$ for 30,000 iterations.

\subsection{Ablation Studies}
\label{SubSecAblation}


\begin{figure*}
    \centering
    \vspace{-5mm}
    \includegraphics[width=\linewidth,trim={0 3.5cm 0 3cm},clip]{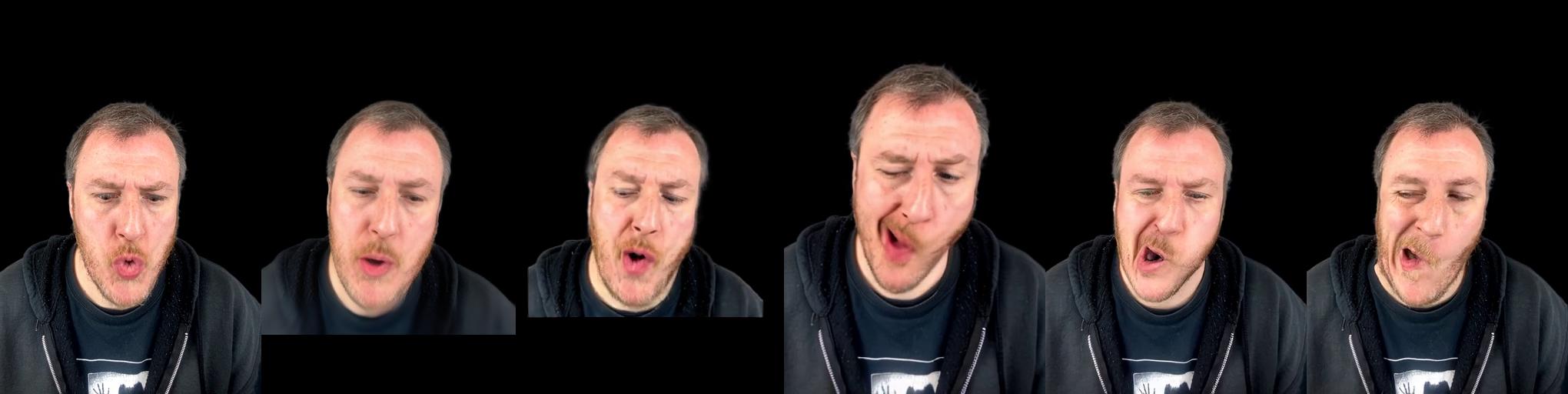}
    \includegraphics[width=\linewidth,trim={0 3cm 0 3.5cm},clip]{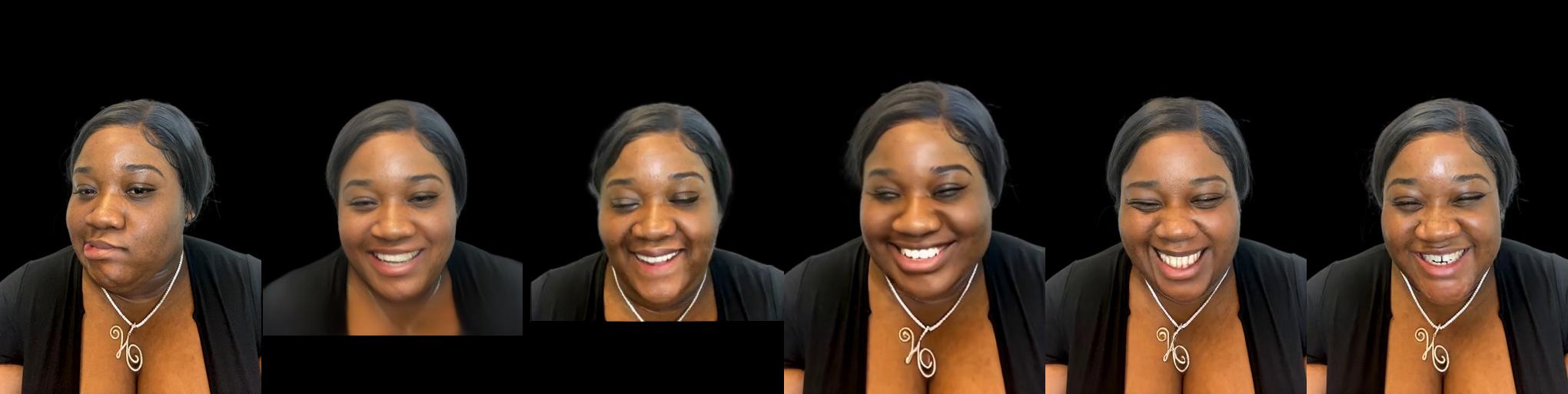}
    \includegraphics[width=\linewidth,trim={0 4.5cm 0 0.5cm},clip]{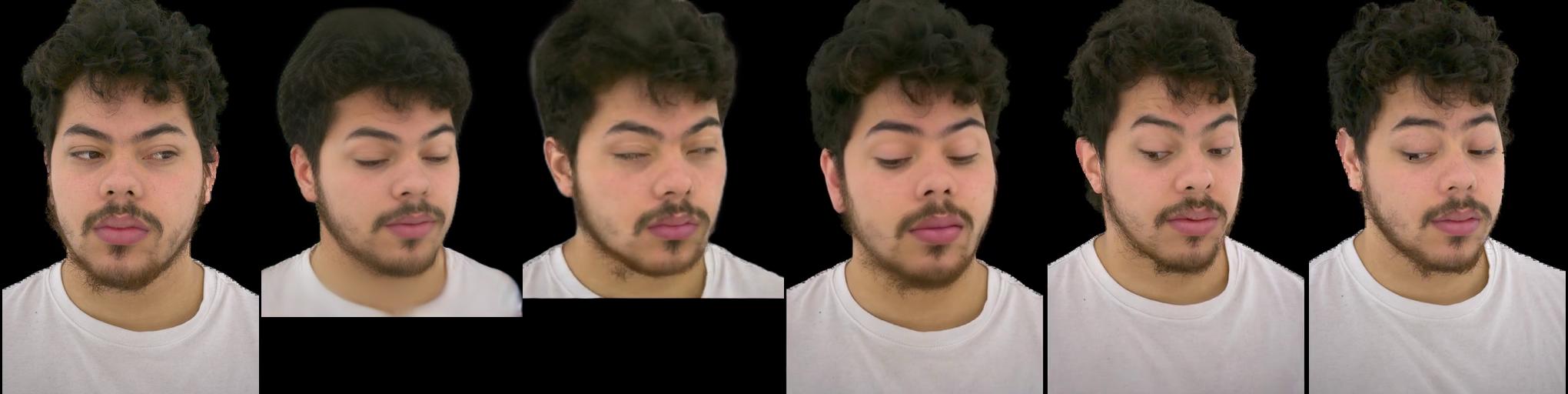}
    \begin{tabular}{p{80pt}p{80pt}p{70pt}p{80pt}p{70pt}p{90pt}}
	 \centering Input & \centering  GAGAvatar~\cite{chu2024generalizable}  & \centering CAP4D~\cite{Taubner2024CAP4DCA}  & \centering HunyuanPortrait~\cite{xu2025hunyuanportrait} & \centering \ Ours 
	 & \centering GT
    \end{tabular}   
    \vspace{-5mm}
    \caption{\textbf{Comparison against state-of-the-arts on the Phone (first two rows) and Studio (bottom row) Dataset}. Previous methods struggle to achieve sufficient photorealism (\eg blurred beard and hair) and precise control on expression or view angle, while our method produces results with sharp details and accurate controls.}
    \label{fig:mgr_col}
    \vspace{-5mm}
\end{figure*}

We conduct comprehensive ablation studies to validate each design of our method. Qualitative and quantitative results are presented in Fig.~\ref{fig:ablation} and Tab.~\ref{tab:ablation}, respectively.

\setlength{\tabcolsep}{1.5pt}
\begin{table}[!bt]
	\renewcommand\arraystretch{1.2}
	\begin{center}
		\begin{tabular}{*{7}{c}}
			\toprule
			   Method & PSNR $\uparrow$ & SSIM $\uparrow$ &  CSIM $\uparrow$ & AED $\downarrow$ & IQA $\uparrow$   & FID-VID $\downarrow$ \\ 
			\midrule
            \midrule

           C1 
                 & \tabsecond{21.63}  & \tabsecond{80.37} &  \tabsecond{78.20}
                 & 0.221 & 57.14  & \tabsecond{40.44} \\

            C2 
                 & 16.62 & 70.32 &  65.16
                 & \tabfirst{0.202} & 56.17 &  50.22 \\

            C3
                 & 19.60  & 77.48  & 70.01 
                 & 0.290 & \tabsecond{57.36} & 53.78 \\
                 
            \midrule
            Full
                 & \tabfirst{22.81} & \tabfirst{83.32}  & \tabfirst{78.82}
                 & \tabsecond{0.212} & \tabfirst{60.08} & \tabfirst{20.68} \\

        \bottomrule
        \end{tabular}
        \caption{\textbf{Quantitative Ablation Studies on the DynamincSweep dataset.} 
        C1: without DynamicSweep during training, C2: without body normal maps as head pose conditions, C3: change expression latents to landmarks as expression conditions.  Overall, these variants would lead to worse image and video quality, lower identity similarity, and less accurate expression control. \tabfirst{Best} and \tabsecond{2nd-best} are highlighted.
        } 
        \vspace{-2mm}
        \label{tab:ablation}
        \end{center}
        \vspace{-7mm}
\end{table}

\noindent \textbf{C1: without dynamic Gaussian avatar renderings during training.}
To enable accurate joint control of camera and expression, we augment training data with videos rendered from animatable Gaussian avatars that exhibit simultaneous camera and expression changes. Without this data, the model fails to generalize to joint control modes, as it cannot extrapolate from training data containing only disjoint camera or expression variations.

\noindent \textbf{C2: without normal maps for body pose control.}
Without it, the generated videos exhibit significant head pose deviations from the ground truth (the 2nd row of Fig.~\ref{fig:ablation}). Since expression codes are disentangled to represent facial expressions only, the model generates ambiguous results with inconsistent head poses, leading to decreased image quality metrics (PSNR, LPIPS, SSIM) as shown in Tab.~\ref{tab:ablation}.

\noindent \textbf{C3: latent expression code vs landmarks.}
We compare the latent expression codes in Sec.~\ref{sec:condition} against 2D facial landmarks. As shown in Fig.~\ref{fig:ablation} and Tab.~\ref{tab:ablation}, landmark conditioning fails to capture subtle facial expressions, resulting in degraded metrics, while the expression latents provide richer representation of fine-grained expression dynamics.



\subsection{Comparisons against State-of-the-arts}

\noindent \textbf{Phone Dataset.}
We compare against prior methods on front-view talking face video generation on the Phone dataset. As shown in Fig.~\ref{fig:mgr_col}, our method can generate complex facial expression and wrinkle details, while prior methods tend to produce results with muted expressions or imprecise pose control. The metric evaluation in Tab.~\ref{tab:comparisons} demonstrated that our method consistently and significantly outperform state-of-the-art methods in almost all metrics.

\setlength{\tabcolsep}{1.5pt}
\begin{table*}[!hbt]
	\renewcommand\arraystretch{0.8}
	\begin{center}
		\begin{tabular}{c|*{9}{c}}
			\toprule
	
			   Dataset & Method & PSNR $\uparrow$ & SSIM $\uparrow$  & LPIPS $\downarrow$ 
               &  CSIM $\uparrow$ & AED $\downarrow$ 
               & IQA $\uparrow$  & FID-VID $\downarrow$  & FVD $\downarrow$  \\ 
			\midrule
            \midrule
           & GAGAvatar~\cite{chu2024generalizable}
                 & \tabsecond{21.45}  & \tabsecond{76.50} & \tabsecond{0.173}
                 & \tabsecond{78.91}  & \tabfirst{0.191} 
                 & 52.50\ & 48.87 & \tabsecond{0.030}  \\

           Phone Capture & CAP4D~\cite{Taubner2024CAP4DCA}
                 & 16.58 & 71.01 & 0.203   
                 & 72.48 & 0.231 
                 & 51.45 & 23.81 & 0.031 \\

           & HunyuanPortrait~\cite{xu2025hunyuanportrait}
                 & 17.18 & 71.37  & 0.216  
                 &  70.91 & 0.199 
                 & \tabsecond{56.58} &  35.37 & 0.032 \\
                 
            \cmidrule{2-10}
            & Ours
                 & \tabfirst{24.68} & \tabfirst{82.85}  & \tabfirst{0.071} & \tabfirst{86.15}  & \tabsecond{0.203} 
                 & \tabfirst{71.16} & 
                 \tabfirst{21.49}  & \tabfirst{0.007} 
                 \\

        \bottomrule

			 
			\midrule
           & GAGAvatar~\cite{chu2024generalizable}
                 & \tabsecond{21.11} & \tabsecond{83.19}  & \tabsecond{0.186}  
                 & \tabsecond{80.11} & \tabsecond{0.156}  
                 & 52.50  & \tabsecond{41.32} & \tabsecond{0.055} \\

           Stuido Capture & CAP4D~\cite{Taubner2024CAP4DCA}  
                 & 14.64 & 74.79 & 0.292  
                 & 77.50 & 0.190 
                 & 49.80 & 44.82 & 0.072 \\

           & HunyuanPortrait~\cite{xu2025hunyuanportrait}
                 & 15.26 & 63.43 & 0.433  
                 & 41.62 & 0.192  
                 & \tabsecond{54.39}  & 83.67 & 0.099  \\
                 
            \cmidrule{2-10}
            & Ours 
                 & \tabfirst{24.45} & \tabfirst{83.80} & \tabfirst{0.118}    & \tabfirst{85.15} 
                 & \tabfirst{0.137} 
                 & \tabfirst{66.81}  & \tabfirst{45.28}  & \tabfirst{0.025} \\

        \bottomrule

			 
            \midrule
           & GAGAvatar~\cite{chu2024generalizable}
                  & \tabsecond{21.11} & \tabsecond{83.19}  & \tabsecond{0.186}  
                  & \tabsecond{80.11} & 0.156  
                  & \tabsecond{52.87}  & 41.32 & 0.055 \\ 
           ViewSweep & CAP4D~\cite{Taubner2024CAP4DCA}
                 & 19.90  & 76.84  & 0.262  
                 & 76.93 & \tabsecond{0.154}  
                 & 50.99 & 96.20 & \tabsecond{0.024} \\

           & HunyuanPortrait~\cite{xu2025hunyuanportrait}
                 & 15.92 & 71.66  & 0.337  
                & 51.59 & 0.175  
                & 26.35  & \tabsecond{40.94}  & 0.215\\

            \cmidrule{2-10}
            & Ours 
                 & \tabfirst{23.25} & \tabfirst{84.55}  & \tabfirst{0.133}  & \tabfirst{81.62} & \tabfirst{0.136} 
                 & \tabfirst{60.77}  & \tabfirst{19.51} &\tabfirst{0.011}  \\

        \bottomrule

			 
            \midrule
           & GAGAvatar~\cite{chu2024generalizable}
                 & \tabsecond{20.58} & \tabsecond{81.93}  & \tabsecond{0.200}  
                 & \tabsecond{77.41} & \tabfirst{0.185} 
                 & \tabsecond{52.50} & 45.32 & 0.041 \\

            DynamicSweep & CAP4D~\cite{Taubner2024CAP4DCA}
                 & 16.05 & 78.41 & 0.260  
                 & 74.00 & 0.214 
                 & 49.97  & \tabsecond{22.39} & \tabsecond{0.032} \\

            & HunyuanPortrait~\cite{xu2025hunyuanportrait}
                 & 15.61 &  70.71  & 0.348  
                 & 52.14 & 0.219   
                 & 32.44  & 41.93  & 0.166 \\
                 
            \cmidrule{2-10}
            & Ours
                 & \tabfirst{22.95} & \tabfirst{83.58}  & \tabfirst{0.137}  & \tabfirst{79.98} & \tabsecond{0.207} 
                 & \tabfirst{61.00} & \tabfirst{20.68}  & \tabfirst{0.008}  \\

        \bottomrule
        \end{tabular}
        \caption{\textbf{ Comparisons against State-of-the-art methods on} Phone Capture, Studio Capture, ViewSweep, DynamicSweep datasets. \tabfirst{Best} and \tabsecond{2nd-best} are highlighted.  
        } 
        \label{tab:comparisons}
        \end{center}
        \vspace{-6mm}
\end{table*}

    

\begin{figure*}
    \centering
    \includegraphics[width=\linewidth,trim={0 2.5cm 0 2.5cm},clip]{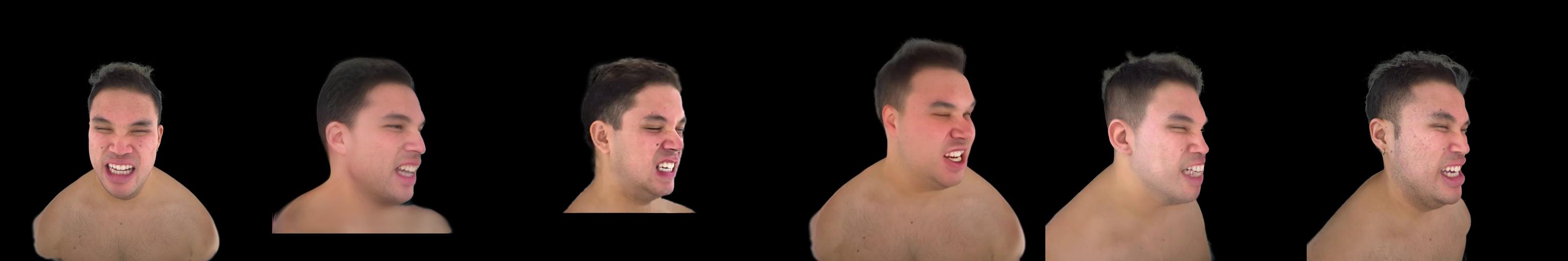}
    \includegraphics[width=\linewidth,trim={0 3.5cm 0 1.5cm},clip]{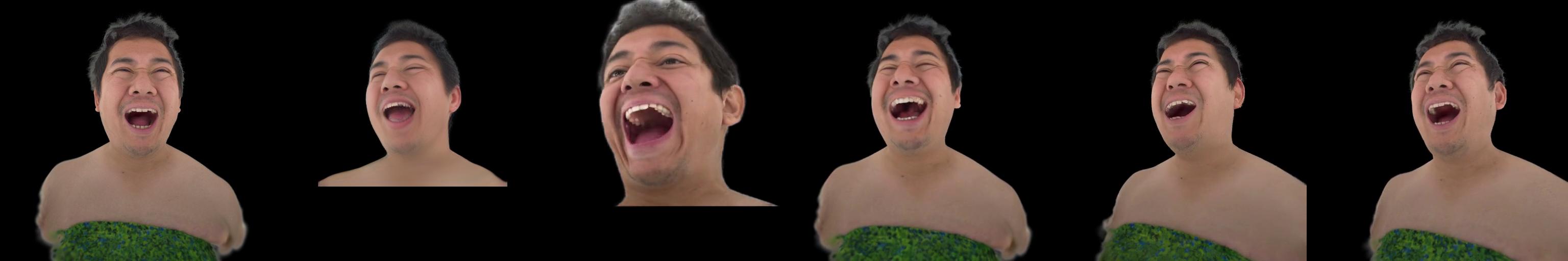}
        \includegraphics[width=\linewidth,trim={0 4.5cm 0 1cm},clip]{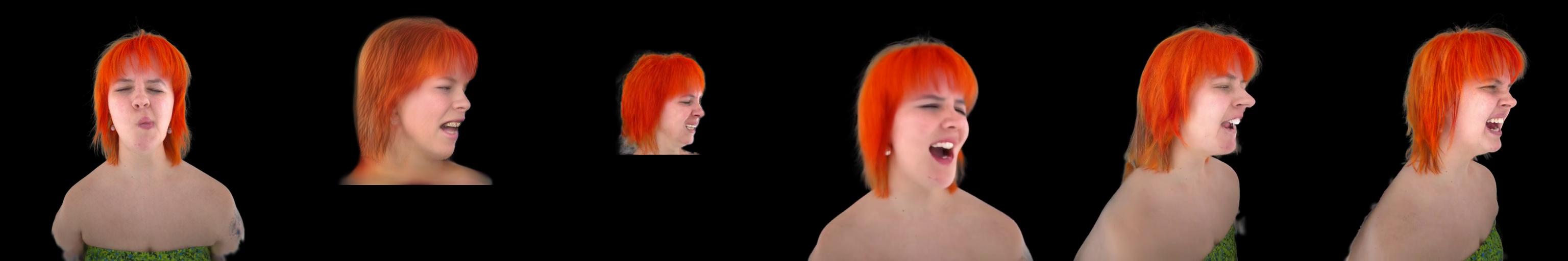}
        \includegraphics[width=\linewidth,trim={0 3cm 0 3cm},clip]{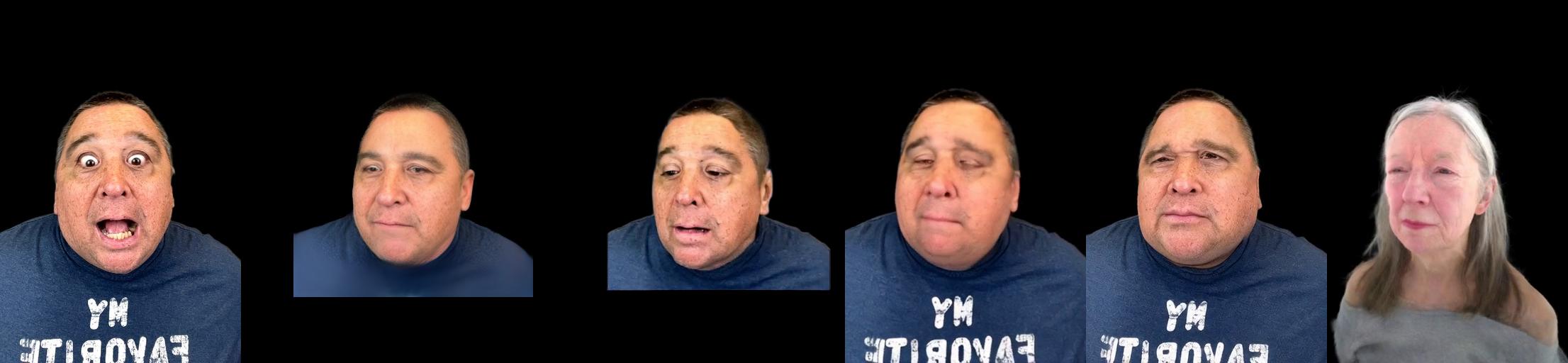}
    \begin{tabular}{p{80pt}p{90pt}p{85pt}p{85pt}p{70pt}p{70pt}}
        \centering Input & \centering  GAGAvatar~\cite{chu2024generalizable}  & \centering CAP4D~\cite{Taubner2024CAP4DCA}  & \centering HunyuanPortrait~\cite{xu2025hunyuanportrait} & \centering \ Ours 
        & \centering GT
    \end{tabular}
    \vspace{-2mm}
    \caption{\textbf{Comparison against state-of-the-arts on the ViewSweep (top two rows) and DynamicSweep (bottom two row) Dataset}, \ie with static expression (ViewSweep) and dynamic expression (DynamicSweep) under random camera trajectories. Noting that \textbf{last row is a cross-reenactment} result. It demonstrates the superiority of our method on identity preservation,  expression transfer and view control.}
    \label{fig:psgsmdtime}
    \vspace{-5mm}
\end{figure*}

\noindent \textbf{Studio Dataset.}
In the Studio dataset, we generate talking face videos under a fixed novel viewpoint. As seen in Fig.~\ref{fig:psgsmdtime}, GAGAvatar and CAP4D struggle to achieve sufficient photorealism, accurate expression control, and fine details such as hair strands. HunyuanPortrait also produces results with a clear deviation from the target view angle. In contrast, our result videos are with better ID preservation and facial expression details such as wrinkles. The superiority of our method is also evidenced by the metrics in Tab.~\ref{tab:comparisons}.

\noindent \textbf{ViewSweep Dataset.}
We evaluate the capability of camera control on the synthetic ViewSweep dataset. Qualitative comparisons in Fig.~\ref{fig:mgr_col} demonstrate that our method preserves identity and expression details, while synthesizes photo-realistc novel views. The improved metrics in Tab.~\ref{tab:comparisons} supports this conclusion as well regarding image quality, identity similarity, and video quality.

\noindent \textbf{DynamicSweep Dataset.}
We also compare the capacity to achieve simultaneous camera and expression control on the DynamicSweep dataset. As seen in Fig.~\ref{fig:psgsmdtime}, our method generates more consistent images in terms of ID and viewpoint for both self driving, and cross-identity animation, which are consistent with metric improvements.

\section{Conclusion}\label{sec:conclusion}
In this work, we propose a controllable portrait video animation method via disentangled conditioning of expression, pose, and camera viewpoint, enabling flexible combinations of different control modes. We leverage synthetic datasets rendered from high-quality animatable Gaussian avatars, to generate videos with static expressions (camera motion only) or time-varying expressions (joint camera and expression dynamics). These synthetic datasets, combined with real-world monocular videos and multi-view videos from professional studios, provide comprehensive supervision for finetuning our video diffusion model. For accurate facial expression control, we employ implicitly defined expression latents to modulate intermediate features via adaptive layer normalization. Extensive experiments demonstrate that our method achieves superior performance in controllable portrait animation with high realism, expressiveness, and view consistency. We hope our work inspires future research on controllable portrait animation for VR/AR applications and beyond.

{
    \small
    \bibliographystyle{ieeenat_fullname}
    \bibliography{main}

\begin{thebibliography}{84}
\providecommand{\natexlab}[1]{#1}
\providecommand{\url}[1]{\texttt{#1}}
\expandafter\ifx\csname urlstyle\endcsname\relax
  \providecommand{\doi}[1]{doi: #1}\else
  \providecommand{\doi}{doi: \begingroup \urlstyle{rm}\Url}\fi

\bibitem[Agrawal et~al.(2025)Agrawal, Akinyemi, Alvero, Behrooz, Buffalini,
  Carlucci, Chen, Chen, Chen, Cheng, et~al.]{seamless_interaction}
Vasu Agrawal, Akinniyi Akinyemi, Kathryn Alvero, Morteza Behrooz, Julia
  Buffalini, Fabio~Maria Carlucci, Joy Chen, Junming Chen, Zhang Chen, Shiyang
  Cheng, et~al.
\newblock Seamless interaction: Dyadic audiovisual motion modeling and
  large-scale dataset.
\newblock \emph{arXiv preprint arXiv:2506.22554}, 2025.

\bibitem[Alexander et~al.(2010)Alexander, Rogers, Lambeth, Chiang, Ma, Wang,
  and Debevec]{alexander2010digital}
Oleg Alexander, Mike Rogers, William Lambeth, Jen-Yuan Chiang, Wan-Chun Ma,
  Chuan-Chang Wang, and Paul Debevec.
\newblock The digital emily project: Achieving a photorealistic digital actor.
\newblock \emph{IEEE Computer Graphics and Applications}, 30\penalty0
  (4):\penalty0 20--31, 2010.

\bibitem[Ba et~al.(2016)Ba, Kiros, and Hinton]{ba2016layer}
Jimmy~Lei Ba, Jamie~Ryan Kiros, and Geoffrey~E Hinton.
\newblock Layer normalization.
\newblock \emph{arXiv preprint arXiv:1607.06450}, 2016.

\bibitem[Bai et~al.(2025)Bai, Xia, Fu, Wang, Mu, Cao, Liu, Hu, Bai, Wan, and
  Zhang]{Bai2025ReCamMasterCG}
Jianhong Bai, Menghan Xia, Xiao Fu, Xintao Wang, Lianrui Mu, Jinwen Cao, Zuozhu
  Liu, Haoji Hu, Xiang Bai, Pengfei Wan, and Di Zhang.
\newblock Recammaster: Camera-controlled generative rendering from a single
  video.
\newblock \emph{arXiv preprint arXiv:2503.11647}, 2025.

\bibitem[Baldridge et~al.(2024)Baldridge, Bauer, Bhutani, Brichtova, Bunner,
  Castrejon, Chan, Chen, Dieleman, Du, et~al.]{baldridge2024imagen}
Jason Baldridge, Jakob Bauer, Mukul Bhutani, Nicole Brichtova, Andrew Bunner,
  Lluis Castrejon, Kelvin Chan, Yichang Chen, Sander Dieleman, Yuqing Du,
  et~al.
\newblock Imagen 3.
\newblock \emph{arXiv preprint arXiv:2408.07009}, 2024.

\bibitem[Behrouzi and Shahroudnejad(2024)]{behrouzi2024maskrenderer}
Tina Behrouzi and Atefeh Shahroudnejad.
\newblock Maskrenderer: 3d-infused multi-mask realistic face reenactment.
\newblock \emph{Pattern Recognition}, 155, 2024.

\bibitem[Blanz and Vetter(1999)]{blanz1999morphable}
Volker Blanz and Thomas Vetter.
\newblock A morphable model for the synthesis of 3d faces.
\newblock In \emph{ACM SIGGRAPH}, pages 187--194, 1999.

\bibitem[Blattmann et~al.(2023{\natexlab{a}})Blattmann, Dockhorn, Kulal,
  Mendelevitch, Kilian, Lorenz, Levi, English, Voleti, Letts,
  et~al.]{blattmann2023stable}
Andreas Blattmann, Tim Dockhorn, Sumith Kulal, Daniel Mendelevitch, Maciej
  Kilian, Dominik Lorenz, Yam Levi, Zion English, Vikram Voleti, Adam Letts,
  et~al.
\newblock Stable video diffusion: Scaling latent video diffusion models to
  large datasets.
\newblock \emph{arXiv preprint arXiv:2311.15127}, 2023{\natexlab{a}}.

\bibitem[Blattmann et~al.(2023{\natexlab{b}})Blattmann, Rombach, Ling,
  Dockhorn, Kim, Fidler, and Kreis]{blattmann2023align}
Andreas Blattmann, Robin Rombach, Huan Ling, Tim Dockhorn, Seung~Wook Kim,
  Sanja Fidler, and Karsten Kreis.
\newblock Align your latents: High-resolution video synthesis with latent
  diffusion models.
\newblock In \emph{IEEE/CVF Conference on Computer Vision and Pattern
  Recognition}, pages 22563--22575, 2023{\natexlab{b}}.

\bibitem[Cao et~al.(2022)Cao, Simon, Kim, Schwartz, Zollhoefer, Saito,
  Lombardi, Wei, Belko, Yu, Sheikh, and Saragih]{cao2022authentic}
Chen Cao, Tomas Simon, Jin~Kyu Kim, Gabe Schwartz, Michael Zollhoefer, Shunsuke
  Saito, Stephen Lombardi, Shih-En Wei, Danielle Belko, Shoou-I Yu, Yaser
  Sheikh, and Jason Saragih.
\newblock Authentic volumetric avatars from a phone scan.
\newblock \emph{ACM Transactions on Graphics}, 41\penalty0 (4), 2022.

\bibitem[Chen et~al.(2024{\natexlab{a}})Chen, Mihajlovic, Wang, Prokudin, and
  Tang]{chen2024morphable}
Xiyi Chen, Marko Mihajlovic, Shaofei Wang, Sergey Prokudin, and Siyu Tang.
\newblock Morphable diffusion: 3d-consistent diffusion for single-image avatar
  creation.
\newblock In \emph{IEEE/CVF Conference on Computer Vision and Pattern
  Recognition}, pages 10359--10370, 2024{\natexlab{a}}.

\bibitem[Chen et~al.(2024{\natexlab{b}})Chen, Jin, Sun, Ni, Zhou, Qin, Chen,
  Huang, and Wang]{chen2024bringyourown}
Zechen Chen, Long Jin, Haolin Sun, Yilun Ni, Yucheng Zhou, Siyu Qin, Yun Chen,
  Haozhe Huang, and Jingdong Wang.
\newblock Bring your own character: A holistic solution for automatic facial
  animation generation of customized characters.
\newblock \emph{arXiv preprint arXiv:2402.13724}, 2024{\natexlab{b}}.

\bibitem[Chen et~al.(2025)Chen, Wang, Wang, Wang, Sun, and Liu]{chen2025v3d}
Zilong Chen, Yikai Wang, Feng Wang, Zhengyi Wang, Fuchun Sun, and Huaping Liu.
\newblock V3d: Video diffusion models are effective 3d generators.
\newblock \emph{IEEE Transactions on Pattern Analysis and Machine
  Intelligence}, pages 1--18, 2025.

\bibitem[Cheng et~al.(2022)Cheng, Cun, Zhang, Xia, Yin, Zhu, Wang, Wang, and
  Wang]{cheng2022videoretalking}
Kun Cheng, Xiaodong Cun, Yong Zhang, Menghan Xia, Fei Yin, Mingrui Zhu, Xuan
  Wang, Jue Wang, and Nannan Wang.
\newblock Videoretalking: Audio-based lip synchronization for talking head
  video editing in the wild.
\newblock In \emph{ACM SIGGRAPH Asia}, pages 1--9, 2022.

\bibitem[Chu and Harada(2024)]{chu2024generalizable}
Xuangeng Chu and Tatsuya Harada.
\newblock Generalizable and animatable gaussian head avatar.
\newblock \emph{Advances in Neural Information Processing Systems},
  37:\penalty0 57642--57670, 2024.

\bibitem[Deng et~al.(2019{\natexlab{a}})Deng, Guo, Niannan, and
  Zafeiriou]{deng2018arcface}
Jiankang Deng, Jia Guo, Xue Niannan, and Stefanos Zafeiriou.
\newblock Arcface: Additive angular margin loss for deep face recognition.
\newblock In \emph{IEEE Conference on Computer Vision and Pattern Recognition},
  2019{\natexlab{a}}.

\bibitem[Deng et~al.(2019{\natexlab{b}})Deng, Yang, Xu, Chen, Jia, and
  Tong]{deng2019accurate}
Yu Deng, Jiaolong Yang, Sicheng Xu, Dong Chen, Yunde Jia, and Xin Tong.
\newblock Accurate 3d face reconstruction with weakly-supervised learning: From
  single image to image set.
\newblock In \emph{IEEE Conference on Computer Vision and Pattern Recognition
  Workshops}, 2019{\natexlab{b}}.

\bibitem[Ding et~al.(2023)Ding, Zhang, Xia, Jebe, Tu, and
  Zhang]{ding2023diffusionrig}
Zheng Ding, Xuaner Zhang, Zhihao Xia, Lars Jebe, Zhuowen Tu, and Xiuming Zhang.
\newblock Diffusionrig: Learning personalized priors for facial appearance
  editing.
\newblock In \emph{IEEE/CVF Conference on Computer Vision and Pattern
  Recognition}, pages 12736--12746, 2023.

\bibitem[Doukas et~al.(2021)Doukas, Zafeiriou, and
  Sharmanska]{doukas2021headgan}
Michail~Christos Doukas, Stefanos Zafeiriou, and Viktoriia Sharmanska.
\newblock Headgan: One-shot neural head synthesis and editing.
\newblock In \emph{IEEE/CVF International conference on Computer Vision}, pages
  14398--14407, 2021.

\bibitem[Drobyshev et~al.(2022)Drobyshev, Chelishev, Khakhulin, Ivakhnenko,
  Lempitsky, and Zakharov]{drobyshev2022megaportraits}
Nikita Drobyshev, Jenya Chelishev, Taras Khakhulin, Aleksei Ivakhnenko, Victor
  Lempitsky, and Egor Zakharov.
\newblock Megaportraits: One-shot megapixel neural head avatars.
\newblock In \emph{ACM International Conference on Multimedia}, pages
  2663--2671, 2022.

\bibitem[Gao et~al.(2024)Gao, Holynski, Henzler, Brussee, Martin-Brualla,
  Srinivasan, Barron, and Poole]{Gao2024CAT3DCA}
Ruiqi Gao, Aleksander Holynski, Philipp Henzler, Arthur Brussee, Ricardo
  Martin-Brualla, Pratul~P. Srinivasan, Jonathan~T. Barron, and Ben Poole.
\newblock Cat3d: Create anything in 3d with multi-view diffusion models.
\newblock \emph{arXiv preprint arXiv:2405.10314}, 2024.

\bibitem[Gao et~al.(2025)Gao, Zhou, Liu, Zhou, and Zhang]{gao2025learn2control}
Xuan Gao, Jingtao Zhou, Dongyu Liu, Yuqi Zhou, and Juyong Zhang.
\newblock Controlling avatar diffusion with learnable gaussian embedding, 2025.

\bibitem[Goodfellow et~al.(2020)Goodfellow, Pouget-Abadie, Mirza, Xu,
  Warde-Farley, Ozair, Courville, and Bengio]{goodfellow2020generative}
Ian Goodfellow, Jean Pouget-Abadie, Mehdi Mirza, Bing Xu, David Warde-Farley,
  Sherjil Ozair, Aaron Courville, and Yoshua Bengio.
\newblock Generative adversarial networks.
\newblock \emph{Communications of the ACM}, 63\penalty0 (11):\penalty0
  139--144, 2020.

\bibitem[Guo et~al.(2024{\natexlab{a}})Guo, Yi, Zhou, Bergman, Lingelbach, and
  Yu]{guo2024real}
Hanzhong Guo, Hongwei Yi, Daquan Zhou, Alexander~William Bergman, Michael
  Lingelbach, and Yizhou Yu.
\newblock Real-time one-step diffusion-based expressive portrait videos
  generation.
\newblock \emph{arXiv preprint arXiv:2412.13479}, 2024{\natexlab{a}}.

\bibitem[Guo et~al.(2024{\natexlab{b}})Guo, Zhang, Liu, Zhong, Zhang, Wan, and
  Zhang]{guo2024liveportrait}
Jianzhu Guo, Dingyun Zhang, Xiaoqiang Liu, Zhizhou Zhong, Yuan Zhang, Pengfei
  Wan, and Di Zhang.
\newblock Liveportrait: Efficient portrait animation with stitching and
  retargeting control.
\newblock \emph{arXiv preprint arXiv:2407.03168}, 2024{\natexlab{b}}.

\bibitem[Guo et~al.(2023)Guo, Yang, Rao, Liang, Wang, Qiao, Agrawala, Lin, and
  Dai]{guo2023animatediff}
Yuwei Guo, Ceyuan Yang, Anyi Rao, Zhengyang Liang, Yaohui Wang, Yu Qiao,
  Maneesh Agrawala, Dahua Lin, and Bo Dai.
\newblock Animatediff: Animate your personalized text-to-image diffusion models
  without specific tuning.
\newblock \emph{arXiv preprint arXiv:2307.04725}, 2023.

\bibitem[Guo et~al.(2024{\natexlab{c}})Guo, Yang, Rao, Liang, Wang, Qiao,
  Agrawala, Lin, and Dai]{guo2024animatediff}
Yuwei Guo, Ceyuan Yang, Anyi Rao, Zhengyang Liang, Yaohui Wang, Yu Qiao,
  Maneesh Agrawala, Dahua Lin, and Bo Dai.
\newblock Animatediff: Animate your personalized text-to-image diffusion models
  without specific tuning.
\newblock In \emph{International Conference on Learning Representations},
  2024{\natexlab{c}}.

\bibitem[He et~al.(2025)He, Xu, Guo, Wetzstein, Dai, Li, and
  Yang]{he2025cameractrl}
Hao He, Yinghao Xu, Yuwei Guo, Gordon Wetzstein, Bo Dai, Hongsheng Li, and
  Ceyuan Yang.
\newblock Cameractrl: Enabling camera control for video diffusion models.
\newblock In \emph{International Conference on Learning Representations}, 2025.

\bibitem[Hu et~al.(2024)Hu, Gao, Zhang, Sun, Zhang, and
  Bo]{hu2024animateanyone}
Li Hu, Xin Gao, Peng Zhang, Ke Sun, Bang Zhang, and Liefeng Bo.
\newblock Animate anyone: Consistent and controllable image-to-video synthesis
  for character animation.
\newblock In \emph{IEEE/CVF Conference on Computer Vision and Pattern
  Recognition}, pages 8153--8163, 2024.

\bibitem[Karras et~al.(2023)Karras, Holynski, Wang, and
  Kemelmacher-Shlizerman]{karras2023dreampose}
Johanna Karras, Aleksander Holynski, Ting-Chun Wang, and Ira
  Kemelmacher-Shlizerman.
\newblock Dreampose: Fashion image-to-video synthesis via stable diffusion.
\newblock In \emph{IEEE/CVF International Conference on Computer Vision}, pages
  22623--22633, 2023.

\bibitem[Khirodkar et~al.(2024)Khirodkar, Bagautdinov, Martinez, Zhaoen, James,
  Selednik, Anderson, and Saito]{khirodkar2024sapiens}
Rawal Khirodkar, Timur Bagautdinov, Julieta Martinez, Su Zhaoen, Austin James,
  Peter Selednik, Stuart Anderson, and Shunsuke Saito.
\newblock Sapiens: Foundation for human vision models.
\newblock In \emph{European Conference on Computer Vision}, pages 206--228.
  Springer, 2024.

\bibitem[Kirschstein et~al.(2023)Kirschstein, Qian, Giebenhain, Walter, and
  Nie{\ss}ner]{kirschstein2023nersemble}
Tobias Kirschstein, Shenhan Qian, Simon Giebenhain, Tim Walter, and Matthias
  Nie{\ss}ner.
\newblock Nersemble: Multi-view radiance field reconstruction of human heads.
\newblock \emph{ACM Transactions on Graphics}, 42\penalty0 (4):\penalty0 1--14,
  2023.

\bibitem[Kirschstein et~al.(2024)Kirschstein, Giebenhain, and
  Nie{\ss}ner]{Kirschstein2023DiffusionAvatarsDD}
Tobias Kirschstein, Simon Giebenhain, and Matthias Nie{\ss}ner.
\newblock Diffusionavatars: Deferred diffusion for high-fidelity 3d head
  avatars.
\newblock In \emph{IEEE/CVF Conference on Computer Vision and Pattern
  Recognition}, pages 5481--5492, 2024.

\bibitem[Kong et~al.(2024)Kong, Tian, Zhang, Min, Dai, Zhou, Xiong, Li, Wu,
  Zhang, et~al.]{kong2024hunyuanvideo}
Weijie Kong, Qi Tian, Zijian Zhang, Rox Min, Zuozhuo Dai, Jin Zhou, Jiangfeng
  Xiong, Xin Li, Bo Wu, Jianwei Zhang, et~al.
\newblock Hunyuanvideo: A systematic framework for large video generative
  models.
\newblock \emph{arXiv preprint arXiv:2412.03603}, 2024.

\bibitem[Li et~al.(2024)Li, Cao, Schwartz, Khirodkar, Richardt, Simon, Sheikh,
  and Saito]{li2024uravatar}
Junxuan Li, Chen Cao, Gabriel Schwartz, Rawal Khirodkar, Christian Richardt,
  Tomas Simon, Yaser Sheikh, and Shunsuke Saito.
\newblock Uravatar: Universal relightable gaussian codec avatars.
\newblock In \emph{ACM SIGGRAPH Asia}, pages 1--11, 2024.

\bibitem[Li et~al.(2017{\natexlab{a}})Li, Bolkart, Black, Li, and
  Romero]{flame2017}
Tianye Li, Timo Bolkart, Michael.~J. Black, Hao Li, and Javier Romero.
\newblock Learning a model of facial shape and expression from {4D} scans.
\newblock \emph{ACM Transactions on Graphics}, 36\penalty0 (6):\penalty0
  194:1--194:17, 2017{\natexlab{a}}.

\bibitem[Li et~al.(2017{\natexlab{b}})Li, Bolkart, Black, Li, and
  Romero]{li2017learning}
Tianye Li, Timo Bolkart, Michael~J Black, Hao Li, and Javier Romero.
\newblock Learning a model of facial shape and expression from 4d scans.
\newblock \emph{ACM Transactions on Graphics}, 36\penalty0 (6):\penalty0
  194--1, 2017{\natexlab{b}}.

\bibitem[Lin et~al.(2022)Lin, Lindell, Chan, and
  Wetzstein]{lin2022ganinversion}
Connor~Z. Lin, David~B. Lindell, Eric~R. Chan, and Gordon Wetzstein.
\newblock 3d gan inversion for controllable portrait image animation.
\newblock In \emph{ECCV Workshop on Learning to Generate 3D Shapes and Scenes},
  2022.

\bibitem[Lipman et~al.(2022)Lipman, Chen, Ben-Hamu, Nickel, and
  Le]{lipman2022flow}
Yaron Lipman, Ricky~TQ Chen, Heli Ben-Hamu, Maximilian Nickel, and Matt Le.
\newblock Flow matching for generative modeling.
\newblock \emph{arXiv preprint arXiv:2210.02747}, 2022.

\bibitem[Liu et~al.(2025)Liu, Deng, Nam, Rong, Pidhorskyi, Li, Saragih,
  Metaxas, and Cao]{liu2025lucas}
Di Liu, Teng Deng, Giljoo Nam, Yu Rong, Stanislav Pidhorskyi, Junxuan Li, Jason
  Saragih, Dimitris~N. Metaxas, and Chen Cao.
\newblock Lucas: Layered universal codec avatars.
\newblock In \emph{IEEE/CVF Conference on Computer Vision and Pattern
  Recognition}, 2025.

\bibitem[Liu et~al.(2023{\natexlab{a}})Liu, Wu, Hoorick, Tokmakov, Zakharov,
  and Vondrick]{Liu2023Zero1to3ZO}
Ruoshi Liu, Rundi Wu, Basile~Van Hoorick, Pavel Tokmakov, Sergey Zakharov, and
  Carl Vondrick.
\newblock Zero-1-to-3: Zero-shot one image to 3d object.
\newblock \emph{IEEE/CVF International Conference on Computer Vision}, pages
  9264--9275, 2023{\natexlab{a}}.

\bibitem[Liu et~al.(2023{\natexlab{b}})Liu, Lin, Zeng, Long, Liu, Komura, and
  Wang]{liu2023syncdreamer}
Yuan Liu, Cheng Lin, Zijiao Zeng, Xiaoxiao Long, Lingjie Liu, Taku Komura, and
  Wenping Wang.
\newblock Syncdreamer: Generating multiview-consistent images from a
  single-view image.
\newblock \emph{arXiv preprint arXiv:2309.03453}, 2023{\natexlab{b}}.

\bibitem[Lombardi et~al.(2021)Lombardi, Simon, Schwartz, Zollhoefer, Sheikh,
  and Saragih]{lombardi2021mixture}
Stephen Lombardi, Tomas Simon, Gabriel Schwartz, Michael Zollhoefer, Yaser
  Sheikh, and Jason Saragih.
\newblock Mixture of volumetric primitives for efficient neural rendering.
\newblock \emph{ACM Transactions on Graphics}, 40\penalty0 (4):\penalty0 1--13,
  2021.

\bibitem[Ma et~al.(2024{\natexlab{a}})Ma, Weng, Shao, and Zhou]{ma2024gaussian}
Shengjie Ma, Yanlin Weng, Tianjia Shao, and Kun Zhou.
\newblock 3d gaussian blendshapes for head avatar animation.
\newblock In \emph{ACM SIGGRAPH}, 2024{\natexlab{a}}.

\bibitem[Ma et~al.(2024{\natexlab{b}})Ma, Liu, Wang, Pan, He, Yuan, Zeng, Cai,
  Shum, Liu, and Shan]{ma2024followyouremoji}
Yue Ma, Hongyu Liu, Hongfa Wang, Heng Pan, Yingqing He, Junkun Yuan, Ailing
  Zeng, Chengfei Cai, Heung-Yeung Shum, Wei Liu, and Ying Shan.
\newblock Follow-your-emoji: Fine-controllable and expressive freestyle
  portrait animation.
\newblock In \emph{arXiv preprint arXiv:2406.01900}, 2024{\natexlab{b}}.

\bibitem[Mark et~al.(2025)Mark, Hu, Xing, and
  Shan]{Mark2025TrajectoryCrafterRC}
YU Mark, Wenbo Hu, Jinbo Xing, and Ying Shan.
\newblock Trajectorycrafter: Redirecting camera trajectory for monocular videos
  via diffusion models.
\newblock \emph{arXiv preprint arXiv:2503.05638}, 2025.

\bibitem[Martinez et~al.(2024)Martinez, Kim, Romero, Bagautdinov, Saito, Yu,
  Anderson, Zollh{\"o}fer, Wang, Bai, et~al.]{martinez2024codec}
Julieta Martinez, Emily Kim, Javier Romero, Timur Bagautdinov, Shunsuke Saito,
  Shoou-I Yu, Stuart Anderson, Michael Zollh{\"o}fer, Te-Li Wang, Shaojie Bai,
  et~al.
\newblock Codec avatar studio: Paired human captures for complete, driveable,
  and generalizable avatars.
\newblock \emph{Advances in Neural Information Processing Systems},
  37:\penalty0 83008--83023, 2024.

\bibitem[Nirkin et~al.(2019)Nirkin, Keller, and Hassner]{nirkin2019fsgan}
Yuval Nirkin, Yosi Keller, and Tal Hassner.
\newblock Fsgan: Subject agnostic face swapping and reenactment.
\newblock In \emph{IEEE/CVF International Conference on Computer Vision}, pages
  7184--7193, 2019.

\bibitem[Ostrek and Thies(2024)]{Ostrek2024StableVP}
Mirela Ostrek and Justus Thies.
\newblock Stable video portraits.
\newblock In \emph{European Conference on Computer Vision}, 2024.

\bibitem[Peebles and Xie(2023)]{peebles2023scalable}
William Peebles and Saining Xie.
\newblock Scalable diffusion models with transformers.
\newblock In \emph{IEEE/CVF Conference on Computer Vision and Pattern
  Recognition}, pages 4195--4205, 2023.

\bibitem[Perez et~al.(2018)Perez, Strub, De~Vries, Dumoulin, and
  Courville]{perez2018film}
Ethan Perez, Florian Strub, Harm De~Vries, Vincent Dumoulin, and Aaron
  Courville.
\newblock Film: Visual reasoning with a general conditioning layer.
\newblock In \emph{Proceedings of the AAAI conference on artificial
  intelligence}, 2018.

\bibitem[Prinzler et~al.(2025)Prinzler, Zakharov, Sklyarova, Kabadayi, and
  Thies]{prinzler2025joker}
Malte Prinzler, Egor Zakharov, Vanessa Sklyarova, Berna Kabadayi, and Justus
  Thies.
\newblock Joker: Conditional 3d head synthesis with extreme facial expressions.
\newblock In \emph{International Conference on 3D Vision (3DV)}, pages
  1583--1593. IEEE, 2025.

\bibitem[Ren et~al.(2025)Ren, Shen, Huang, Ling, Lu, Nimier-David, Muller,
  Keller, Fidler, and Gao]{Ren2025Gen3C3W}
Xuanchi Ren, Tianchang Shen, Jiahui Huang, Huan Ling, Yifan Lu, Merlin
  Nimier-David, Thomas Muller, Alexander Keller, Sanja Fidler, and Jun Gao.
\newblock Gen3c: 3d-informed world-consistent video generation with precise
  camera control.
\newblock \emph{IEEE/CVF Conference on Computer Vision and Pattern
  Recognition}, pages 6121--6132, 2025.

\bibitem[Rombach et~al.(2022)Rombach, Blattmann, Lorenz, Esser, and
  Ommer]{rombach2022high}
Robin Rombach, Andreas Blattmann, Dominik Lorenz, Patrick Esser, and Bj{\"o}rn
  Ommer.
\newblock High-resolution image synthesis with latent diffusion models.
\newblock In \emph{IEEE/CVF Conference on Computer Vision and Pattern
  Recognition}, pages 10684--10695, 2022.

\bibitem[Schroff et~al.(2015)Schroff, Kalenichenko, and
  Philbin]{schroff2015facenet}
Florian Schroff, Dmitry Kalenichenko, and James Philbin.
\newblock Facenet: A unified embedding for face recognition and clustering.
\newblock In \emph{IEEE Conference on Computer Vision and Pattern Recognition},
  2015.

\bibitem[Shi et~al.(2023{\natexlab{a}})Shi, Chen, Zhang, Liu, Xu, Wei, Chen,
  Zeng, and Su]{Shi2023Zero123AS}
Ruoxi Shi, Hansheng Chen, Zhuoyang Zhang, Minghua Liu, Chao Xu, Xinyue Wei,
  Linghao Chen, Chong Zeng, and Hao Su.
\newblock Zero123++: a single image to consistent multi-view diffusion base
  model.
\newblock \emph{arXiv preprint arXiv:2310.15110}, 2023{\natexlab{a}}.

\bibitem[Shi et~al.(2023{\natexlab{b}})Shi, Wang, Ye, Long, Li, and
  Yang]{Shi2023MVDreamMD}
Yichun Shi, Peng Wang, Jianglong Ye, Mai Long, Kejie Li, and X. Yang.
\newblock Mvdream: Multi-view diffusion for 3d generation.
\newblock \emph{arXiv preprint arXiv:2308.16512}, 2023{\natexlab{b}}.

\bibitem[Siarohin et~al.(2019{\natexlab{a}})Siarohin, Lathuili{\`e}re,
  Tulyakov, Ricci, and Sebe]{siarohin2019first}
Aliaksandr Siarohin, St{\'e}phane Lathuili{\`e}re, Sergey Tulyakov, Elisa
  Ricci, and Nicu Sebe.
\newblock First order motion model for image animation.
\newblock \emph{Advances in Neural Information Processing Systems}, 32,
  2019{\natexlab{a}}.

\bibitem[Siarohin et~al.(2019{\natexlab{b}})Siarohin, Lathuilière, Tulyakov,
  Ricci, and Sebe]{siarohin2019fomm}
Aliaksandr Siarohin, Stéphane Lathuilière, Sergey Tulyakov, Elisa Ricci, and
  Nicu Sebe.
\newblock First order motion model for image animation.
\newblock \emph{Advances in Neural Information Processing Systems},
  2019{\natexlab{b}}.

\bibitem[Tang et~al.(2025)Tang, Davoli, Kirschstein, Schoneveld, and
  Niessner]{tang2025gaf}
Jiapeng Tang, Davide Davoli, Tobias Kirschstein, Liam Schoneveld, and Matthias
  Niessner.
\newblock Gaf: Gaussian avatar reconstruction from monocular videos via
  multi-view diffusion.
\newblock In \emph{IEEE Conference on Computer Vision and Pattern Recognition},
  pages 5546--5558, 2025.

\bibitem[Taubner et~al.(2024)Taubner, Zhang, Tuli, and
  Lindell]{Taubner2024CAP4DCA}
Felix Taubner, Ruihang Zhang, Mathieu Tuli, and David~B. Lindell.
\newblock Cap4d: Creating animatable 4d portrait avatars with morphable
  multi-view diffusion models.
\newblock \emph{IEEE/CVF Conference on Computer Vision and Pattern
  Recognition}, pages 5318--5330, 2024.

\bibitem[Tian et~al.(2024)Tian, Wang, Zhang, and Bo]{tian2024emo}
Linrui Tian, Qi Wang, Bang Zhang, and Liefeng Bo.
\newblock Emo: Emote portrait alive - generating expressive portrait videos
  with audio2video diffusion model under weak conditions.
\newblock \emph{arXiv preprint arXiv:2402.17485}, 2024.

\bibitem[Tong et~al.(2024)Tong, Li, Chen, Wu, and Zhou]{tong2024musepose}
Zhengyan Tong, Chao Li, Zhaokang Chen, Bin Wu, and Wenjiang Zhou.
\newblock Musepose: a pose-driven image-to-video framework for virtual human
  generation.
\newblock \emph{arXiv preprint arXiv:2405.17827}, 2024.

\bibitem[Unterthiner et~al.(2018)Unterthiner, van Steenkiste, Kurach, Marinier,
  Michalski, and Gelly]{DBLP:journals/corr/abs-1812-01717}
Thomas Unterthiner, Sjoerd van Steenkiste, Karol Kurach, Rapha{\"{e}}l
  Marinier, Marcin Michalski, and Sylvain Gelly.
\newblock Towards accurate generative models of video: {A} new metric {\&}
  challenges.
\newblock \emph{CoRR}, abs/1812.01717, 2018.

\bibitem[Ververas and Zafeiriou(2022)]{ververas2022f3agan}
Evangelos Ververas and Stefanos Zafeiriou.
\newblock F3a-gan: Facial flow for face animation with generative adversarial
  networks.
\newblock \emph{IEEE Transactions on Circuits and Systems for Video
  Technology}, 2022.

\bibitem[Wan et~al.(2025)Wan, Wang, Ai, Wen, Mao, Xie, Chen, Yu, Zhao, Yang,
  et~al.]{wan2025}
Team Wan, Ang Wang, Baole Ai, Bin Wen, Chaojie Mao, Chen-Wei Xie, Di Chen,
  Feiwu Yu, Haiming Zhao, Jianxiao Yang, et~al.
\newblock Wan: Open and advanced large-scale video generative models.
\newblock \emph{arXiv preprint arXiv:2503.20314}, 2025.

\bibitem[Wang and Shi(2023)]{Wang2023ImageDreamIM}
Peng Wang and Yichun Shi.
\newblock Imagedream: Image-prompt multi-view diffusion for 3d generation.
\newblock \emph{arXiv preprint arXiv:2312.02201}, 2023.

\bibitem[Wang et~al.(2024{\natexlab{a}})Wang, Yu, Zheng, Zhou, and
  Huang]{wang2024vividpose}
Qilin Wang, Qingyuan Yu, Xiaoyu Zheng, Yuan Zhou, and Shuai Huang.
\newblock Vividpose: Advancing stable video diffusion for realistic human image
  animation.
\newblock \emph{arXiv preprint arXiv:2405.18156}, 2024{\natexlab{a}}.

\bibitem[Wang et~al.(2024{\natexlab{b}})Wang, Li, Lin, Zhai, Lin, Yang, Zhang,
  Liu, and Wang]{wang2024disco}
Tan Wang, Linjie Li, Kevin Lin, Yuanhao Zhai, Chung-Ching Lin, Zhengyuan Yang,
  Hanwang Zhang, Zicheng Liu, and Lijuan Wang.
\newblock Disco: Disentangled control for realistic human dance generation.
\newblock In \emph{IEEE/CVF Conference on Computer Vision and Pattern
  Recognition}, 2024{\natexlab{b}}.

\bibitem[Wang et~al.(2021)Wang, Mallya, and Liu]{wang2021facevid2vid}
Ting-Chun Wang, Arun Mallya, and Ming-Yu Liu.
\newblock One-shot free-view neural talking-head synthesis for video
  conferencing.
\newblock In \emph{IEEE/CVF Conference on Computer Vision and Pattern
  Recognition}, 2021.

\bibitem[Wang et~al.(2024{\natexlab{c}})Wang, Yuan, Zhang, Chen, Wang, Zhang,
  Shen, Zhao, and Zhou]{wang2024unianimate}
Xiang Wang, Hangjie Yuan, Shiwei Zhang, Dayou Chen, Jiuniu Wang, Yingya Zhang,
  Yujun Shen, Deli Zhao, and Jingren Zhou.
\newblock Unianimate: Taming unified video diffusion models for consistent
  human image animation.
\newblock \emph{arXiv preprint arXiv:2406.01188}, 2024{\natexlab{c}}.

\bibitem[Wei et~al.(2024)Wei, Yang, and Wang]{wei2024aniportrait}
Huawei Wei, Zejun Yang, and Zhisheng Wang.
\newblock Aniportrait: Audio-driven synthesis of photorealistic portrait
  animation.
\newblock \emph{arXiv preprint arXiv:2403.17694}, 2024.

\bibitem[Wu et~al.(2024)Wu, Gao, Poole, Trevithick, Zheng, Barron, and
  Holynski]{Wu2024CAT4DCA}
Rundi Wu, Ruiqi Gao, Ben Poole, Alex Trevithick, Changxi Zheng, Jonathan~T.
  Barron, and Aleksander Holynski.
\newblock Cat4d: Create anything in 4d with multi-view video diffusion models.
\newblock \emph{IEEE/CVF Conference on Computer Vision and Pattern
  Recognition}, pages 26057--26068, 2024.

\bibitem[Xie et~al.(2024)Xie, Xu, Song, Wang, Shi, and Luo]{xie2024x}
You Xie, Hongyi Xu, Guoxian Song, Chao Wang, Yichun Shi, and Linjie Luo.
\newblock X-portrait: Expressive portrait animation with hierarchical motion
  attention.
\newblock In \emph{ACM SIGGRAPH 2024 Conference Papers}, pages 1--11, 2024.

\bibitem[Xu et~al.(2024)Xu, Zhang, Liew, Yan, Liu, Zhang, Feng, and
  Shou]{xu2024magicanimate}
Zhongcong Xu, Jianfeng Zhang, Jun~Hao Liew, Hanshu Yan, Jia-Wei Liu, Chenxu
  Zhang, Jiashi Feng, and Mike~Zheng Shou.
\newblock Magicanimate: Temporally consistent human image animation using
  diffusion model.
\newblock In \emph{IEEE/CVF Conference on Computer Vision and Pattern
  Recognition}, pages 16016--16025, 2024.

\bibitem[Xu et~al.(2025)Xu, Yu, Zhou, Zhou, Jin, Hong, Ji, Zhu, Cai, Tang,
  et~al.]{xu2025hunyuanportrait}
Zunnan Xu, Zhentao Yu, Zixiang Zhou, Jun Zhou, Xiaoyu Jin, Fa-Ting Hong,
  Xiaozhong Ji, Junwei Zhu, Chengfei Cai, Shiyu Tang, et~al.
\newblock Hunyuanportrait: Implicit condition control for enhanced portrait
  animation.
\newblock In \emph{IEEE/CVF Conference on Computer Vision and Pattern
  Recognition}, pages 15909--15919, 2025.

\bibitem[Yang et~al.(2024{\natexlab{a}})Yang, Zhang, Tang, Qian, and
  Yang]{Yang2024ConsistentAvatarLT}
Haijie Yang, Zhenyu Zhang, Hao Tang, Jianjun Qian, and Jian Yang.
\newblock Consistentavatar: Learning to diffuse fully consistent talking head
  avatar with temporal guidance.
\newblock In \emph{ACM Conference on Multimedia}, 2024{\natexlab{a}}.

\bibitem[Yang et~al.(2024{\natexlab{b}})Yang, Li, Wu, Jing, Li, Ji, Liang, and
  Fan]{yang2024megactor}
Shurong Yang, Huadong Li, Juhao Wu, Minhao Jing, Linze Li, Renhe Ji, Jiajun
  Liang, and Haoqiang Fan.
\newblock Megactor: Harness the power of raw video for vivid portrait
  animation.
\newblock \emph{arXiv preprint arXiv:2405.20851}, 2024{\natexlab{b}}.

\bibitem[Yu et~al.(2024)Yu, Xing, Yuan, Hu, Li, Huang, Gao, Wong, Shan, and
  Tian]{Yu2024ViewCrafterTV}
Wangbo Yu, Jinbo Xing, Li Yuan, Wenbo Hu, Xiaoyu Li, Zhipeng Huang, Xiangjun
  Gao, Tien-Tsin Wong, Ying Shan, and Yonghong Tian.
\newblock Viewcrafter: Taming video diffusion models for high-fidelity novel
  view synthesis.
\newblock \emph{IEEE Transactions on Pattern Analysis and Machine
  Intelligence}, pages 1--18, 2024.

\bibitem[Zeng et~al.(2023)Zeng, Liu, Gao, Liu, Li, Liu, and
  Zhang]{Zeng2023FaceAW}
Bo-Wen Zeng, Xuhui Liu, Sicheng Gao, Boyu Liu, Hong Li, Jianzhuang Liu, and
  Baochang Zhang.
\newblock Face animation with an attribute-guided diffusion model.
\newblock \emph{IEEE/CVF Conference on Computer Vision and Pattern Recognition
  Workshops}, pages 628--637, 2023.

\bibitem[Zhang et~al.(2023)Zhang, Qi, Zhang, Zhang, Wu, Chen, Chen, Wang, and
  Wen]{zhang2023metaportrait}
Bowen Zhang, Chenyang Qi, Pan Zhang, Bo Zhang, HsiangTao Wu, Dong Chen, Qifeng
  Chen, Yong Wang, and Fang Wen.
\newblock Metaportrait: Identity-preserving talking head generation with fast
  personalized adaptation.
\newblock In \emph{IEEE/CVF Conference on Computer Vision and Pattern
  Recognition}, pages 22096--22105, 2023.

\bibitem[Zhi et~al.(2025)Zhi, Li, Liao, Yang, Sun, Chang, Cun, Feng, and
  Han]{zhi2025mv}
Yihao Zhi, Chenghong Li, Hongjie Liao, Xihe Yang, Zhengwentai Sun, Jiahao
  Chang, Xiaodong Cun, Wensen Feng, and Xiaoguang Han.
\newblock Mv-performer: Taming video diffusion model for faithful and
  synchronized multi-view performer synthesis.
\newblock \emph{arXiv preprint arXiv:2510.07190}, 2025.

\bibitem[Zhou et~al.(2025)Zhou, Gao, Voleti, Vasishta, Yao, Boss, Torr,
  Rupprecht, and Jampani]{Zhou2025StableVC}
Jensen Zhou, Hang Gao, Vikram~S. Voleti, Aaryaman Vasishta, Chun-Han Yao, Mark
  Boss, Philip Torr, Christian Rupprecht, and Varun Jampani.
\newblock Stable virtual camera: Generative view synthesis with diffusion
  models.
\newblock \emph{arXiv preprint arXiv:2503.14489}, 2025.

\bibitem[Zhu et~al.(2024)Zhu, Chen, Dai, Xu, Cao, Yao, Zhu, and
  Zhu]{zhu2024champ}
Shenhao Zhu, Junming~Leo Chen, Zuozhuo Dai, Yinghui Xu, Xun Cao, Yao Yao, Hao
  Zhu, and Siyu Zhu.
\newblock Champ: Controllable and consistent human image animation with 3d
  parametric guidance.
\newblock In \emph{European Conference on Computer Vision}, 2024.

\end{thebibliography}
}

\clearpage
\maketitlesupplementary


In this supplementary material, we provide additional details about our dataset curation strategy in Sec.~\ref{SecDataset}. We then introduce further implementation details of our method variants in Sec.~\ref{SecImple}. Next, we present additional results on the Phone Capture, Studio Capture, ViewSweep, and DynamicSweep datasets, including both self-reenactment and cross-reenactment, in Sec.~\ref{SecAddRes}. Finally, we discuss the limitations of our current work and outline future directions in Sec.~\ref{SecLimit}.
We encourage readers to visit the webpage self-contained in our supplementary material for more video generation results.

\section{Dataset Curation}
\label{SecDataset}

A summary of the dataset for disentangled dynamics is described in Tab.~\ref{tab:dataset}, with details explained below.

\subsection{Phone Capture}

\setlength{\tabcolsep}{3pt}
\begin{table*}[!tbp]
	\renewcommand\arraystretch{1.2}
	\begin{center}
        \begin{tabular}{*{8}{c}}
			\toprule
            \multirow{2}{*}{Dataset} & \multirow{2}{*}{Cameras} & \multirow{2}{*}{Identities} & \multirow{2}*{Frames/ID} & \multirow{2}*{Resolution} & \multicolumn{3}{c}{Each video exhibits} \\
            & & & & & View change & Expression change & Pose change \\
			\midrule
            PhoneCapture
                &  1 & 11,976 & 2,000 & 1440x1080 & \xmark & \cmark & \cmark \\ 
                
            StudioCapture
                & 78 & 612 & 4,000 & 2048x1334 &\xmark & \cmark & \cmark\\ 

            ViewSweep
                & Random  & 802 & 12,800 & 1024x1024 & \cmark & \xmark & \xmark \\ 
            
            DynamicSweep
                & Random  & 802 & 4,096 & 1024x1024 & \cmark & \cmark & \cmark \\ 
        
        \bottomrule
        \end{tabular}
        \caption{\textbf{Dataset Overview.} PhoneCapture and StudioCapture datasets contain real video recordings. ViewSweep and DynamicSweep datasets consist of synthetic video renderings based on fitted Gaussian avatars.}
        \label{tab:dataset}
        \end{center}
        \vspace{-4mm}
\end{table*}
We utilize a monocular iPhone video dataset comprising 11,976 unique identities, with each identity contains an average of 4,000 frames from 30 videos, at a resolution of 1440x1080 pixels. For each identity, the video sequences include a variety of actions such as head rotation, brief expressions, and speech. Since phone captures features a static camera set up and focus on facial dynamics across a large number of identities, we primarily leverage the rich identities and diverse facial dynamics in this dataset.

\subsection{Studio Capture}

The entire studio dataset comprises 1414 identities from 78 synchronized cameras at a resolution of 2048x1334 pixels. Each identity is captured with approximately 4,000 frames, encompassing a wide range of facial expressions, head movement, emotional displays, gaze motion, sentence reading, speech, and free face activities. For training efficiency, each identity randomly samples 11 views from the 78 cameras for each capture, ensuring coverage across the entire camera views while reducing computational overhead. We retain the raw captures for 612 identities, for learning the expression dynamics and novel view synthesis.

\subsection{ViewSweep}
The rendering for fitted Gaussian Avatars simulates mobile-like captures by setting up a series of handheld cameras and introducing randomness for each. For each capture, we render 128 distinct camera trajectories, \ie, 100-frame video sequences, with frozen expression and head pose for each trajectory. 
There are two kinds of camera trajectories, spin and spiral. Cameras in spin videos follow an oval path, distance between 25 centimeters and 40 centimeters, and at most 5 elevation degree randomness. Spiral camera trajectory is drawn by fitting a spiral curve based on 4 randomly sampled seed locations, within yaw angle in $[-90^\circ, 90^\circ]$. Camera intrinsics follows an Field of View of $72^\circ$ horizontally and vertically, to simulate the iPhone-like captures.

\subsection{DynamicSweep}
In this setting, instead of maintaining a frozen expression during camera spin or spiral, facial expressions and poses are taken from a random segment (128-frame) of the original studio capture. We render 32 distinct trajectories for each identity, \ie, 128-frame video sequences, among which 16 are camera spin and rest 16 are camera spiral. Camera configurations are adjusted similarly as the ViewSweep dataset. 

\section{Implementation Details}
\label{SecImple}

\subsection{Expression Condition}
We adopt a pretrained imitator face representation~\cite{seamless_interaction} for expression condition. Specifically, the pretrained expression encoder, \ie a typical ResNet34 backbone with a linear head, extracts the 128-dim latent feature from a roll-normalized facial image crop. An alignment encoder, with the same ResNet34 plus linear layer architecture, produces 3D translations for the head and body, and rotation angles for the head alone, from an upper body crop image. These two encoders were trained with a decoder end-to-end for talking head video generation. 

\subsection{Ablation Studies}

Our final model uses a sequence of expression codes as condition to control the faciail expression. For head pose control, we concatenate noisy latents with normal maps rendered from the body mesh tracking as inputs, which are fed into diffusion transformer.
Our model was trained on a combination of Phone, Studio, ViewSweep, and DynamicSweep datasets. To study the effectiveness of each design, we individually remove each design.

\noindent \textbf{Without dynamic Gaussian avatar renderings.}
To enable accurate joint control of camera and expression, we use DynamicSweep dataset, ~\ie videos rendered from animatable Gaussian avatars exhibiting simultaneous camera and expression changes. Without the DynamicSweep dataset, the data ratio for training stages 3 and 4 is as follows: Phone (20\%), Studio (40\%), and ViewSweep (40\%).

\noindent \textbf{Without normal maps for head pose control.}
To assess the impact of normal maps on head pose control, we conduct an ablation study by removing the normal maps rendered from body mesh tracking in the condition fusion layer.

\noindent \textbf{Latent expression code vs. 2D facial landmarks.}
Instead of using latent expression codes for expression control, an alternative approach is facial landmarks detected from the driving video. Specifically, we detect 238 facial landmarks per frame and represent them as 2D point clouds. These landmark sequences are encoded using a transformer with alternating spatial and temporal attention layers, capturing both intra-frame spatial relationships and inter-frame temporal dynamics. Specifically, the landmark encoder is composed of 2 spatial attention and 2 temporal attention layers. To inject the encoded landmark features into the diffusion transformer, we add additional cross-attention layers, enabling the model to incorperate the landmark features at each frame. The training strategy and configuration remain identical to those used for our final model.

\section{Additional Results}
\label{SecAddRes}

\subsection{Self-Reenactment}
\begin{figure*}
    \centering
    \vspace{-5mm}
    \includegraphics[width=\linewidth,trim={0 2.5cm 0 2.5cm},clip]{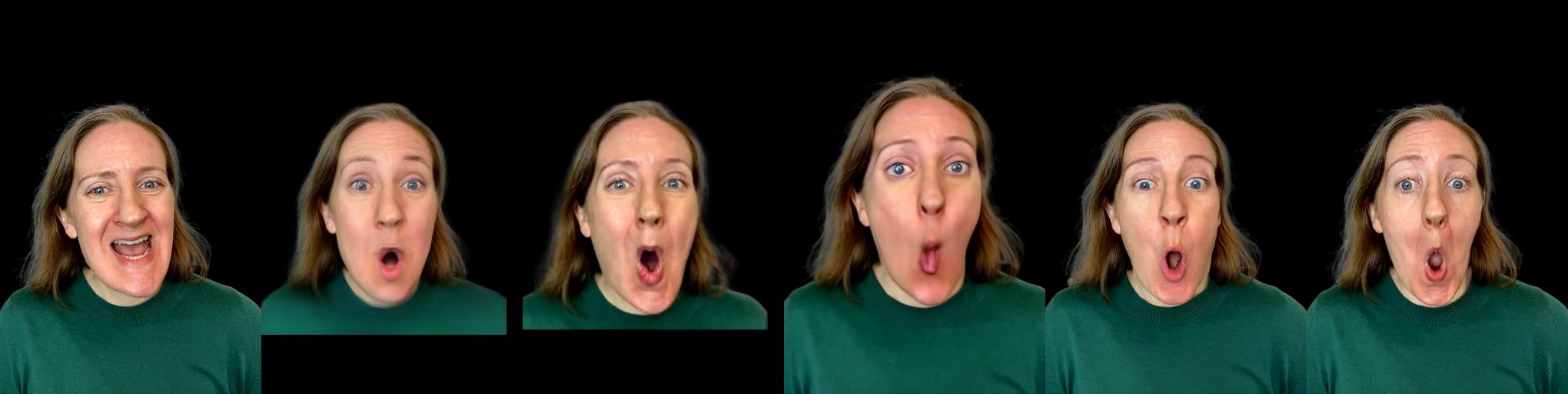}
    \includegraphics[width=\linewidth,trim={0 2.5cm 0 2.5cm},clip]{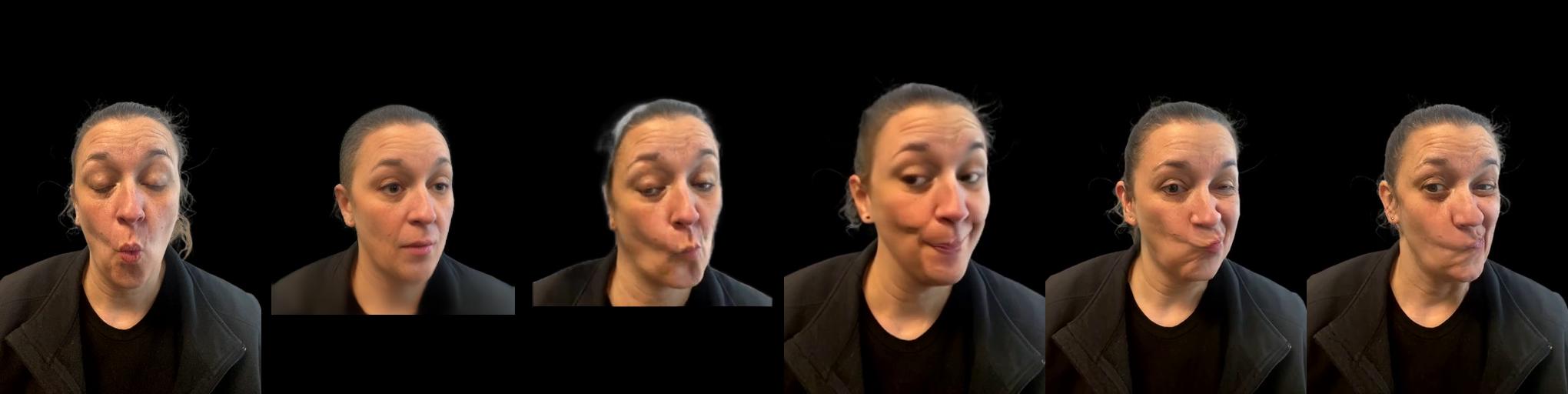}
    \includegraphics[width=\linewidth,trim={0 2.5cm 0 2.5cm},clip]{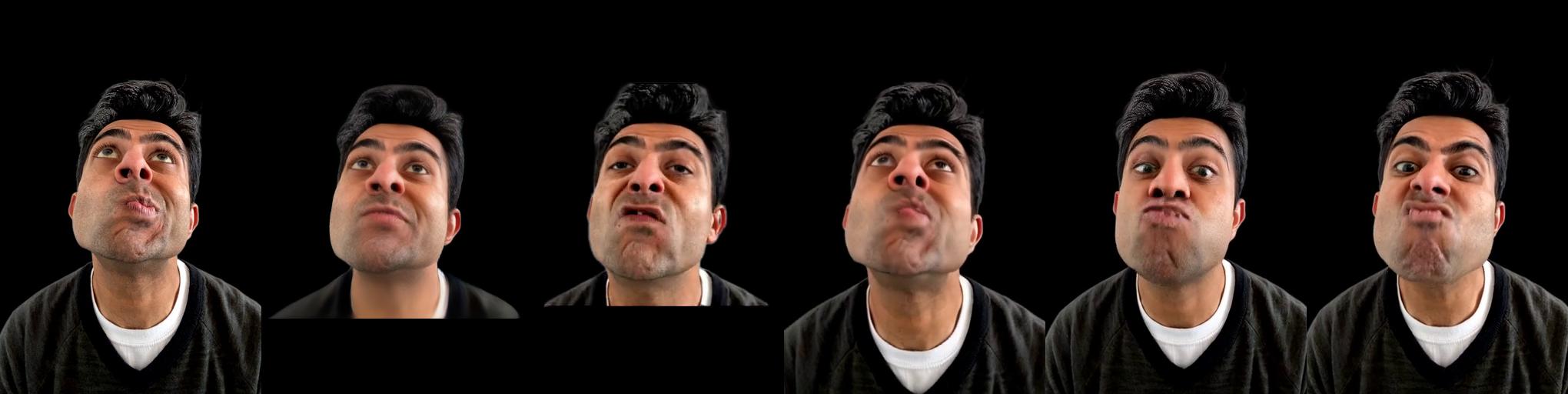}
    \includegraphics[width=\linewidth,trim={0 2.5cm 0 2.5cm},clip]{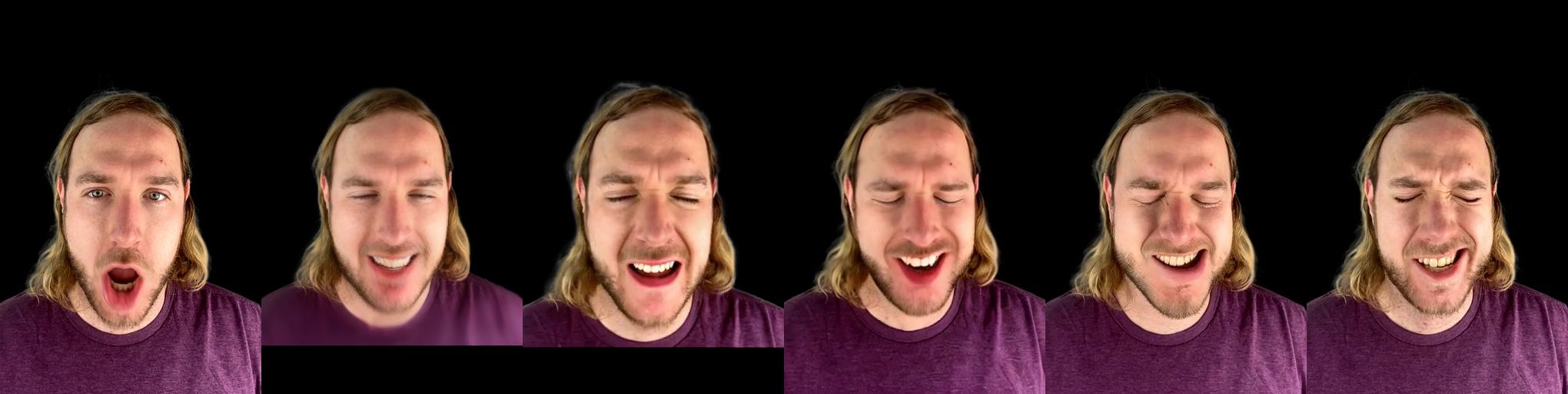}
    \includegraphics[width=\linewidth,trim={0 2.5cm 0 2.5cm},clip]{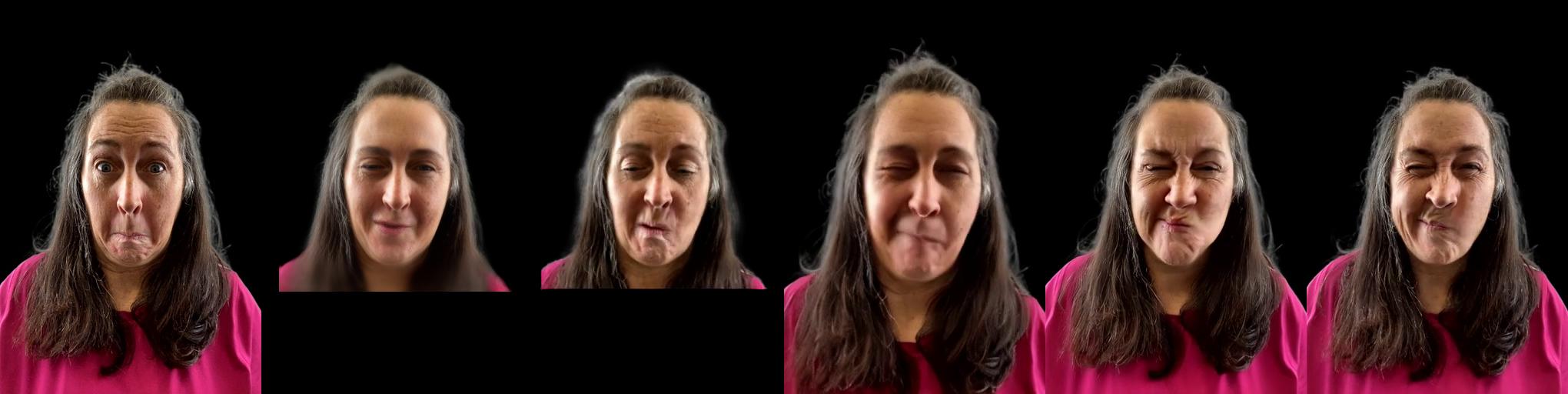}
    \includegraphics[width=\linewidth,trim={0 2.5cm 0 2.5cm},clip]{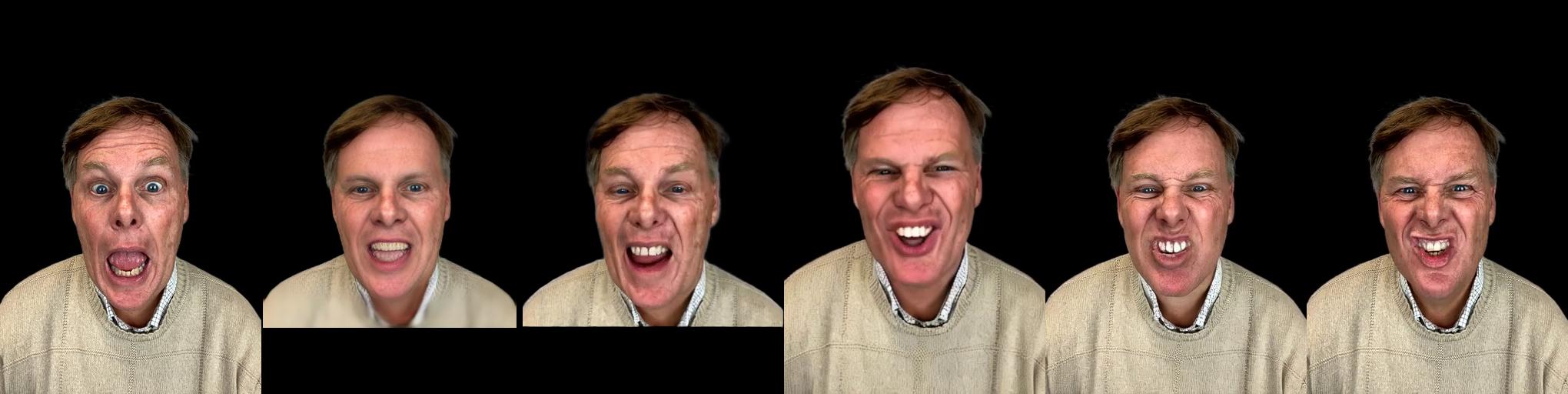}
    \begin{tabular}{p{80pt}p{80pt}p{70pt}p{80pt}p{70pt}p{90pt}}
        \centering Input & \centering  GAGAvatar~\cite{chu2024generalizable}  & \centering CAP4D~\cite{Taubner2024CAP4DCA}  & \centering HunyuanPortrait~\cite{xu2025hunyuanportrait} & \centering \ Ours 
        & \centering GT
    \end{tabular}   
    \vspace{-5mm}
    \caption{\textbf{Comparison against state-of-the-art methods on the Phone Dataset}. The target output is a static, frontal view video with changing expressions. Our method achieves more accurate control over complex facial expressions and additionally enables the generation of hair and torso regions.}
    \label{fig:mgr_supple}
    \vspace{-5mm}
\end{figure*}

\begin{figure*}
    \centering
    \vspace{-5mm}
    \includegraphics[width=\linewidth,trim={0 2.5cm 0 2.5cm},clip]{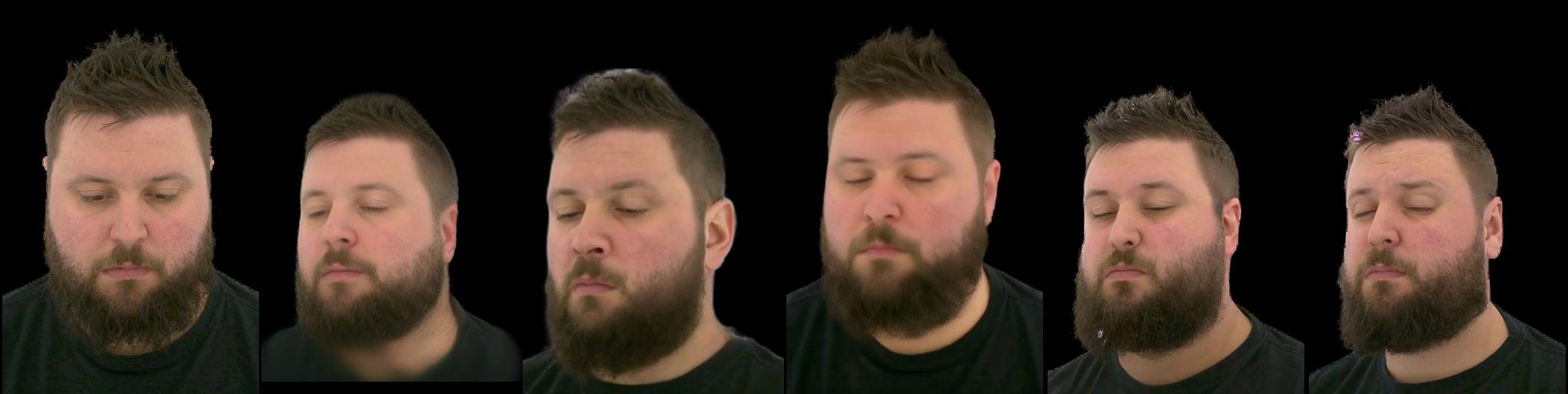}
    \includegraphics[width=\linewidth,trim={0 2.5cm 0 2.5cm},clip]{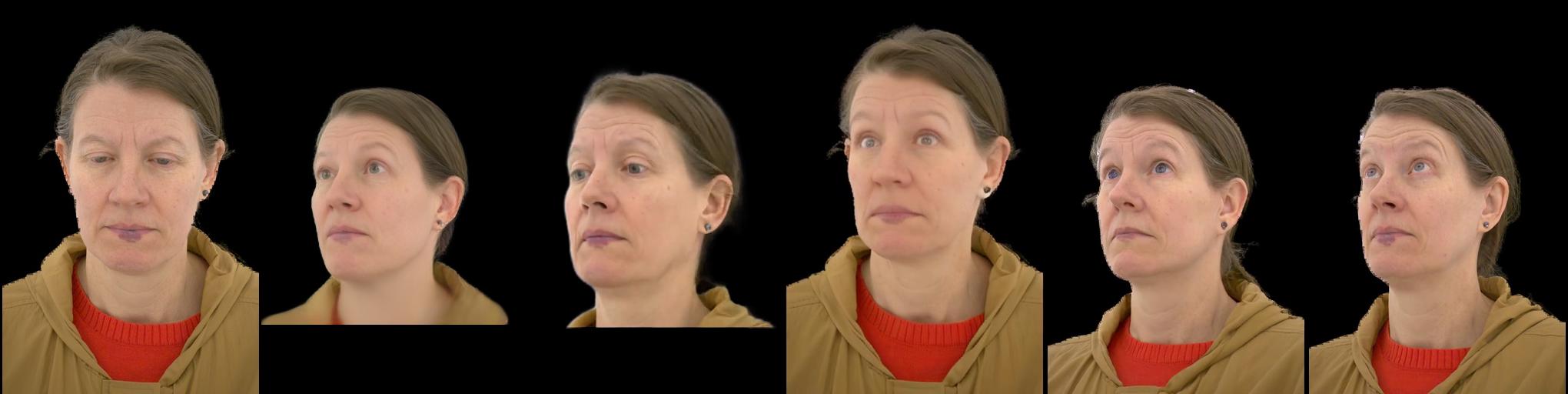}
    \includegraphics[width=\linewidth,trim={0 2.5cm 0 2.5cm},clip]{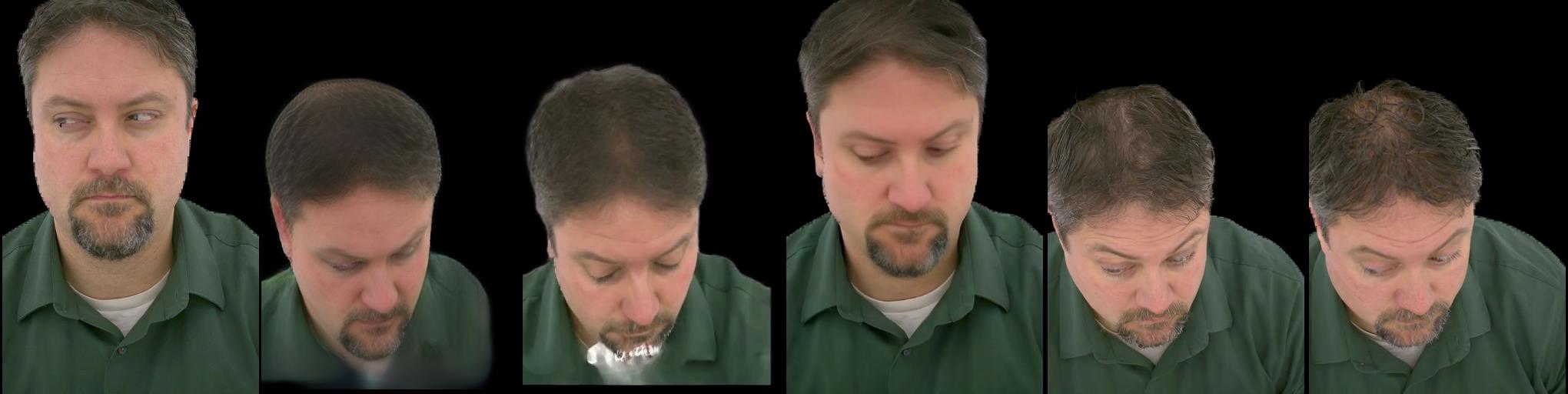}
    \includegraphics[width=\linewidth,trim={0 2.5cm 0 2.5cm},clip]{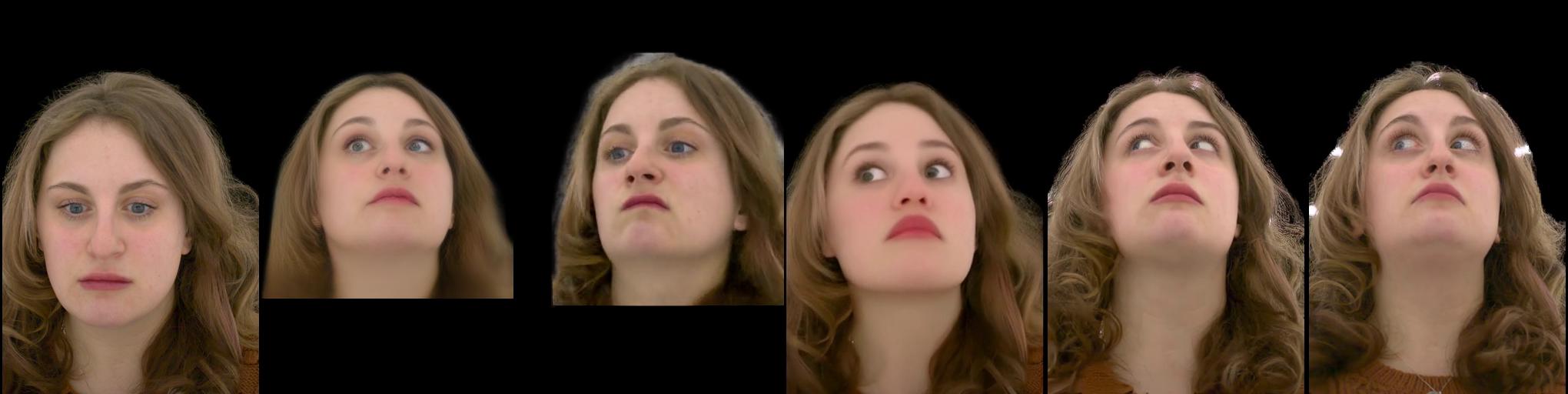}
    \includegraphics[width=\linewidth,trim={0 2.5cm 0 2.5cm},clip]{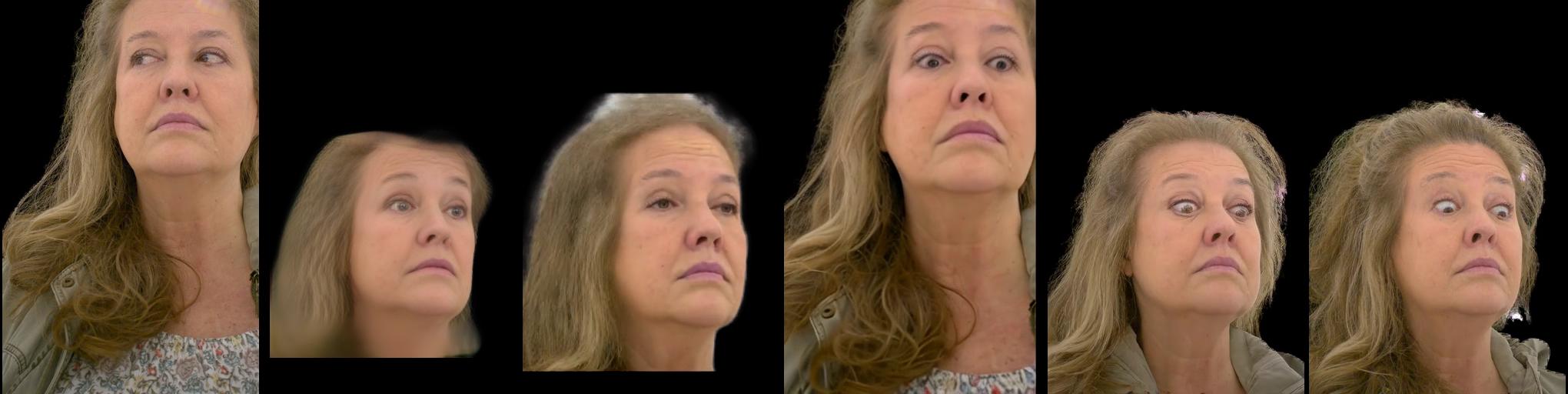}
    \includegraphics[width=\linewidth,trim={0 2.5cm 0 2.5cm},clip]{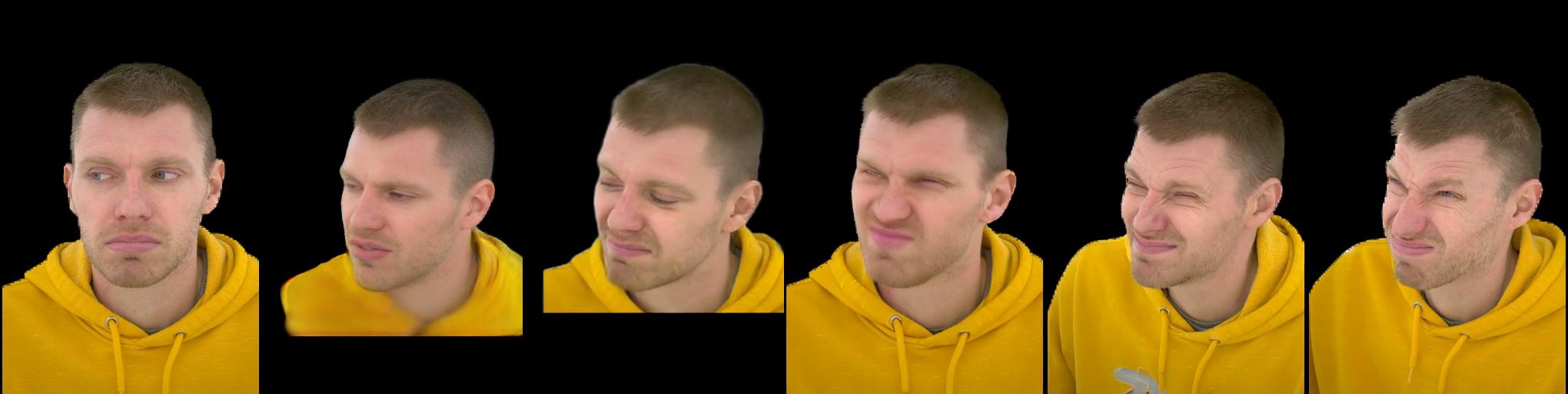}
    \begin{tabular}{p{80pt}p{80pt}p{70pt}p{80pt}p{70pt}p{90pt}}
    \centering Input & \centering  GAGAvatar~\cite{chu2024generalizable}  & \centering CAP4D~\cite{Taubner2024CAP4DCA}  & \centering HunyuanPortrait~\cite{xu2025hunyuanportrait} & \centering \ Ours 
        & \centering GT
    \end{tabular}   
    \vspace{-5mm}
    \caption{\textbf{Comparison against state-of-the-art method on the Studio Dataset}. The desired output is a static, novel view video with changing expressions. Our method generates more plausible eye movements and gaze directions. Compared to HunyuanPortrait, our approach enables more accurate viewpoint control. In contrast to GAGAvatar, we can synthesize hair and beards with sharper details. }
    \label{fig:col_supple}
    \vspace{-5mm}
\end{figure*}

\begin{figure*}
    \centering
    \includegraphics[width=\linewidth,trim={0 1.5cm 0 1.5cm},clip]{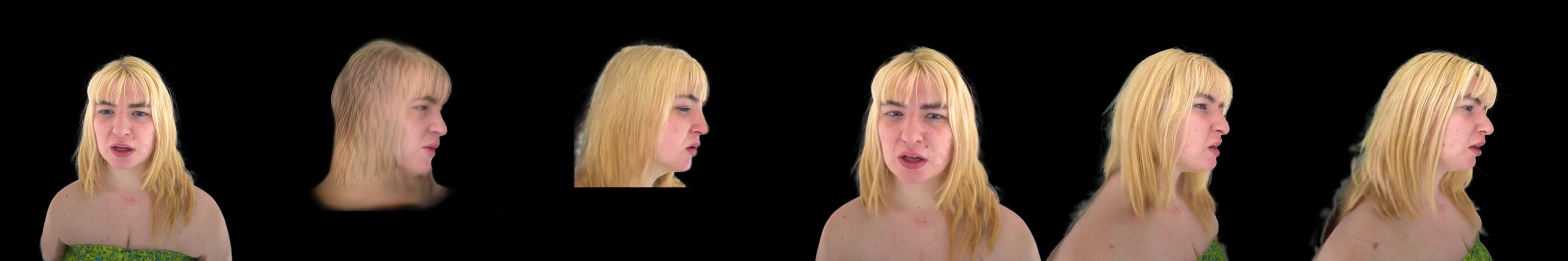}
   \includegraphics[width=\linewidth,trim={0 1.5cm 0 1.5cm},clip]{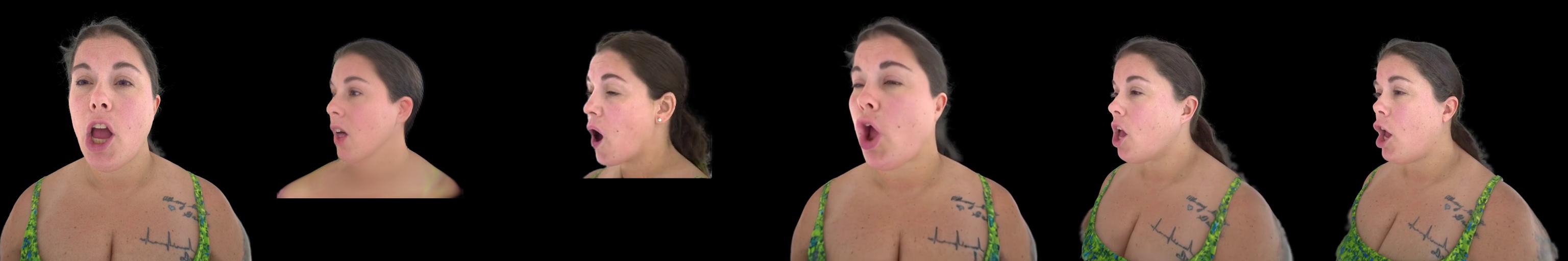}
   \includegraphics[width=\linewidth,trim={0 2.5cm 0 0.5cm},clip]{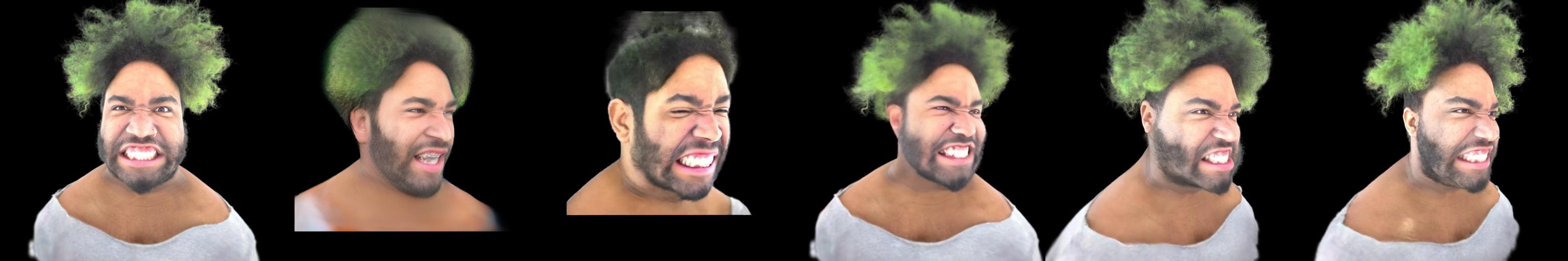}
    \includegraphics[width=\linewidth,trim={0 2.5cm 0 0.5cm},clip]{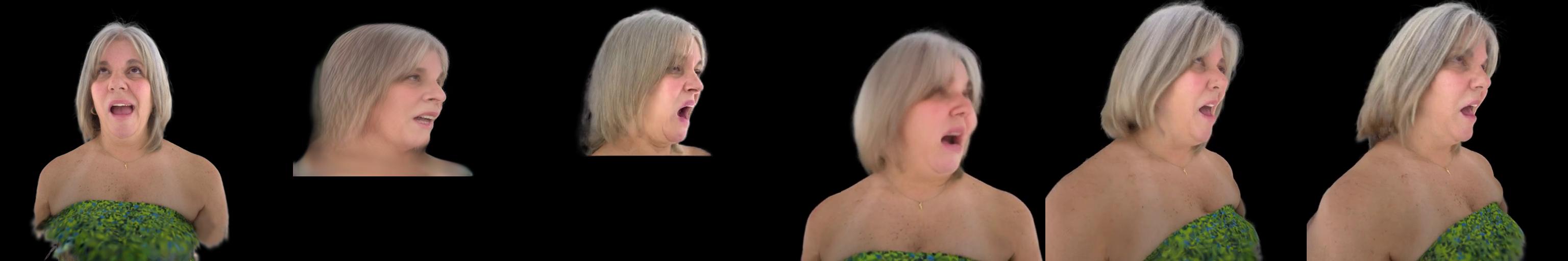}
    \includegraphics[width=\linewidth,trim={0 2.5cm 0 0.5cm},clip]{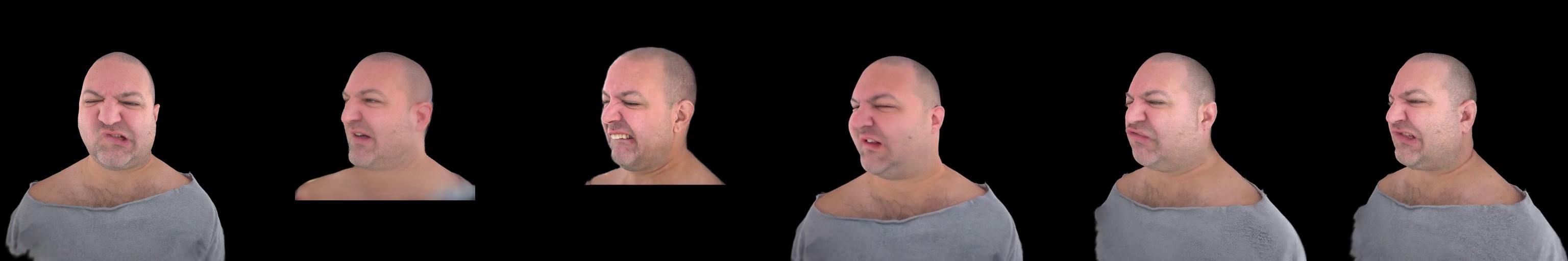}
    \includegraphics[width=\linewidth,trim={0 2.5cm 0 0.5cm},clip]{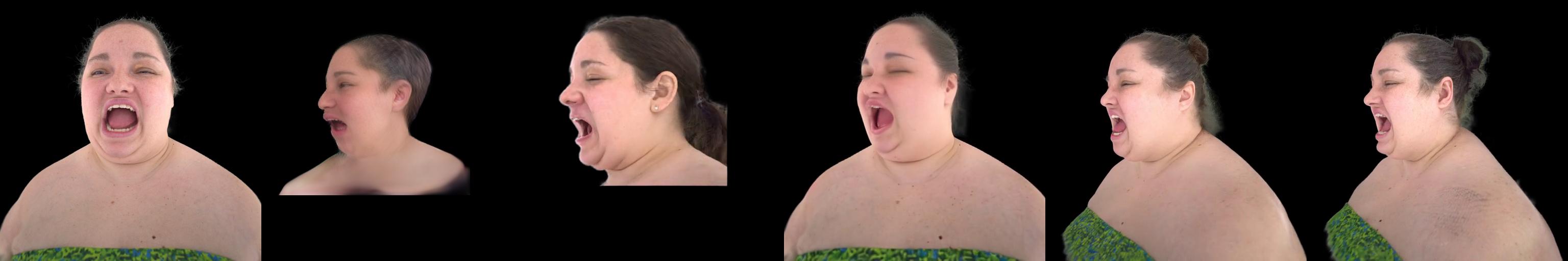}
    \includegraphics[width=\linewidth,trim={0 1.5cm 0 1.5cm},clip]{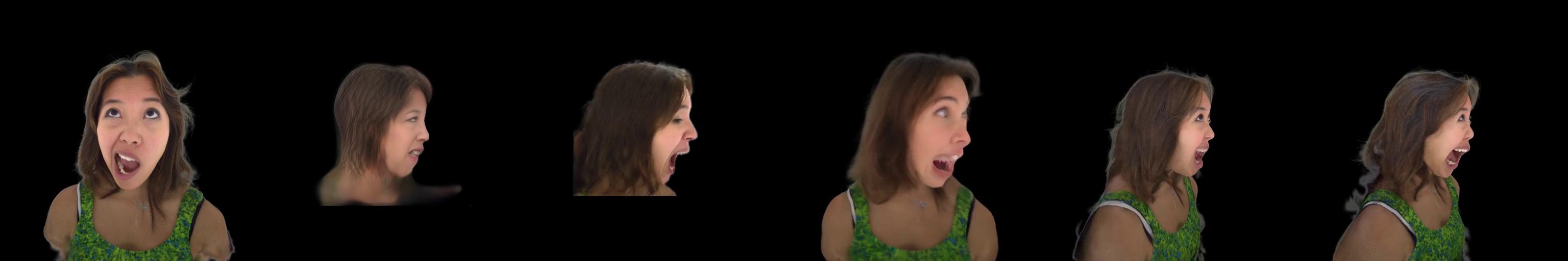}
    \includegraphics[width=\linewidth,trim={0 1.5cm 0 1.5cm},clip]{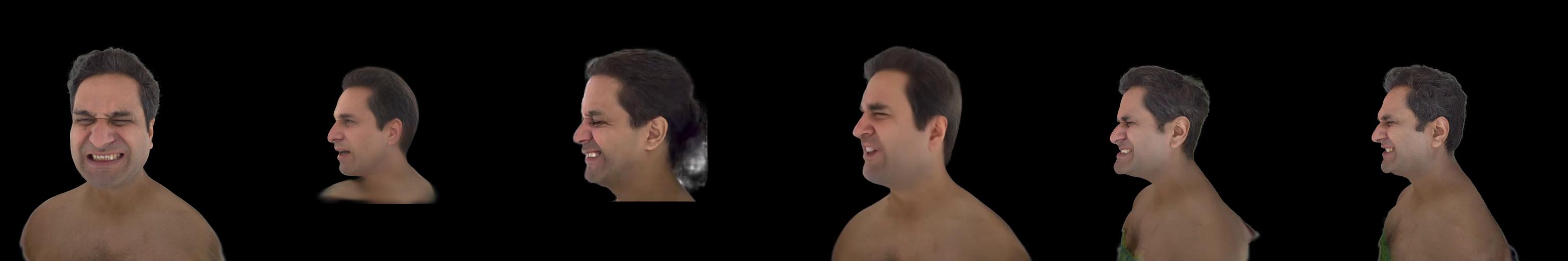}
    \begin{tabular}{p{80pt}p{80pt}p{70pt}p{80pt}p{70pt}p{90pt}}
        \centering Input & \centering  GAGAvatar~\cite{chu2024generalizable}  & \centering CAP4D~\cite{Taubner2024CAP4DCA}  & \centering HunyuanPortrait~\cite{xu2025hunyuanportrait} & \centering \ Ours 
        & \centering GT
    \end{tabular}   
    \caption{\textbf{Comparison against state-of-the-arts on the ViewSweep Dataset}. The desired output is a stack of novel-view images along a camera trajectory, with static expressions. Our method achieves more favorable view synthesis, better preserving facial identity and capturing detailed hair and mouth interior features under hold-out viewpoints. Compared to the recent video diffusion method HunyuanPortrait, our approach enables more precise viewpoint control.}
    \label{fig:viewsweep_supple}
\end{figure*}

\begin{figure*}
    \centering
    \vspace{-5mm}
   \includegraphics[width=\linewidth,trim={0 1.5cm 0 1.5cm},clip]{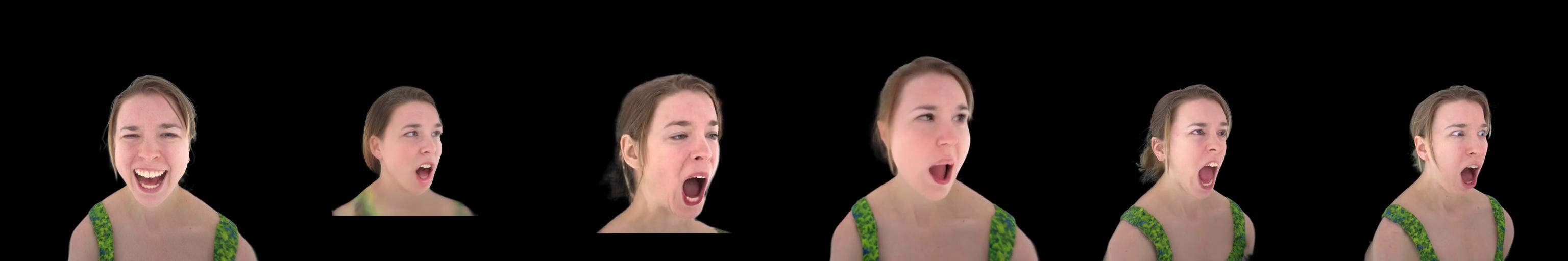}
   \includegraphics[width=\linewidth,trim={0 1.5cm 0 1.5cm},clip]{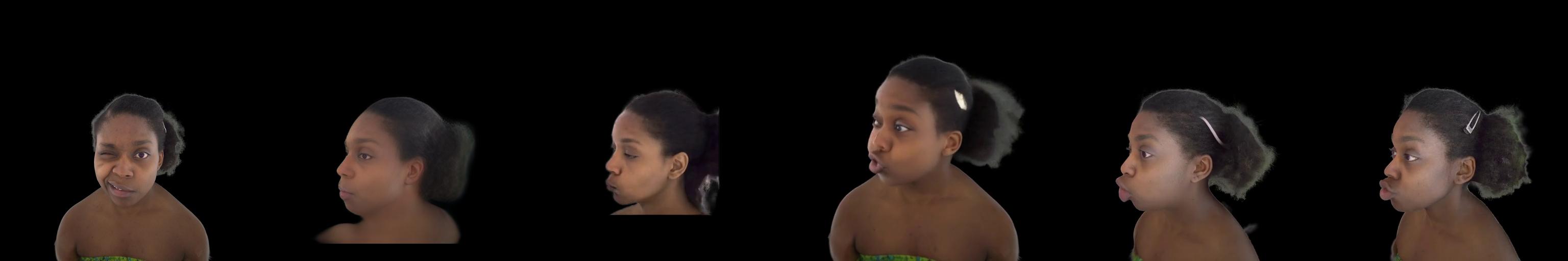}
   \includegraphics[width=\linewidth,trim={0 1.5cm 0 1.5cm},clip]{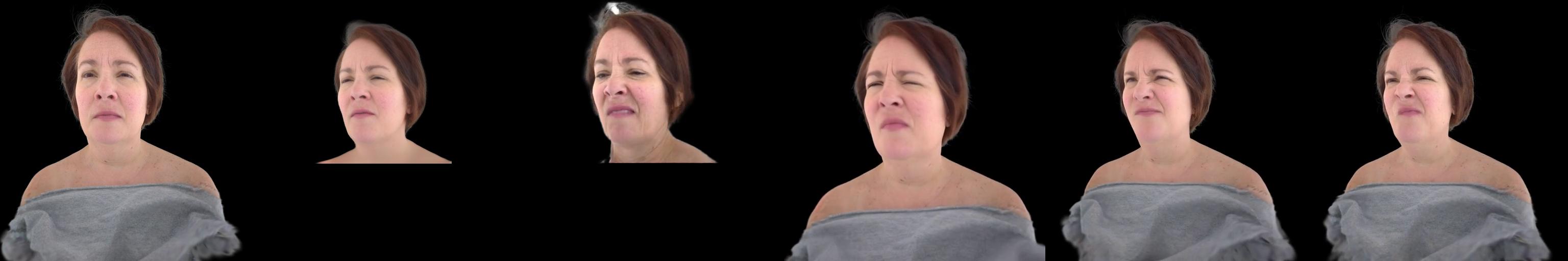}
   \includegraphics[width=\linewidth,trim={0 1.5cm 0 1.5cm},clip]{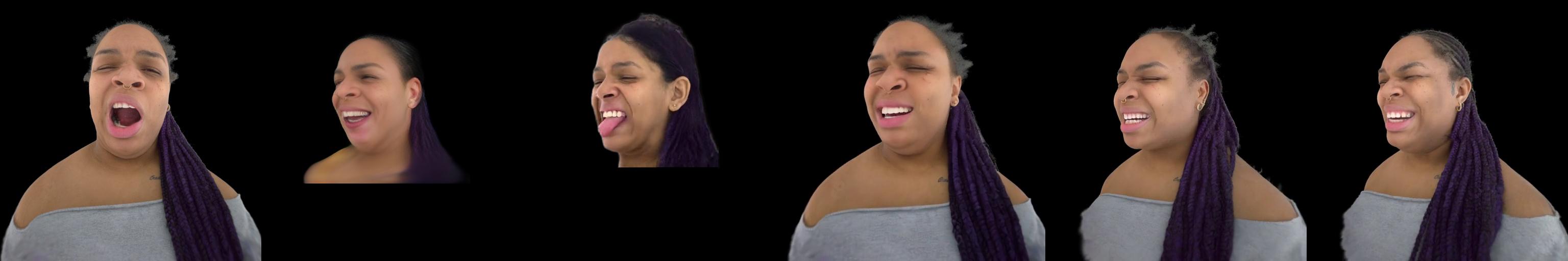}
   \includegraphics[width=\linewidth,trim={0 1.5cm 0 1.5cm},clip]{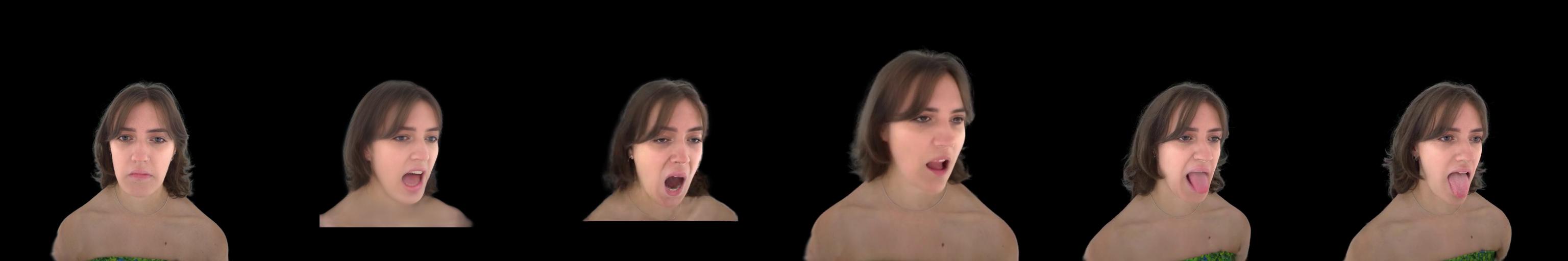}
   \includegraphics[width=\linewidth,trim={0 1.5cm 0 1.5cm},clip]{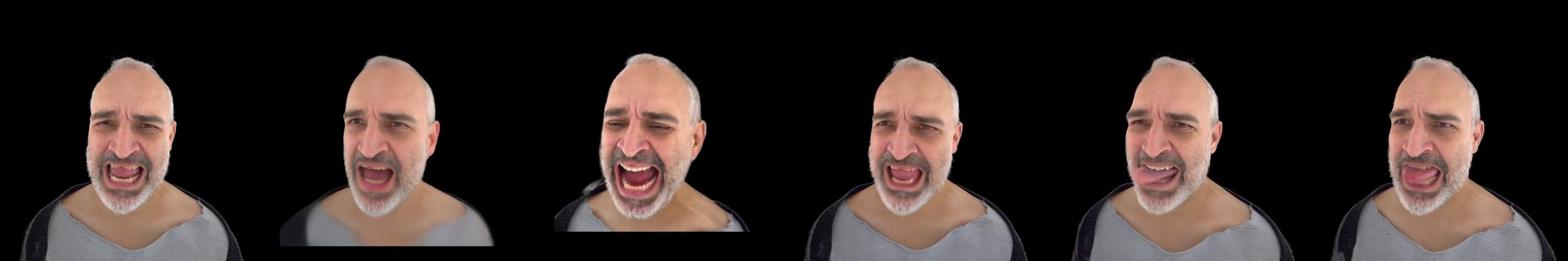}
    \includegraphics[width=\linewidth,trim={0 1.5cm 0 1.5cm},clip]{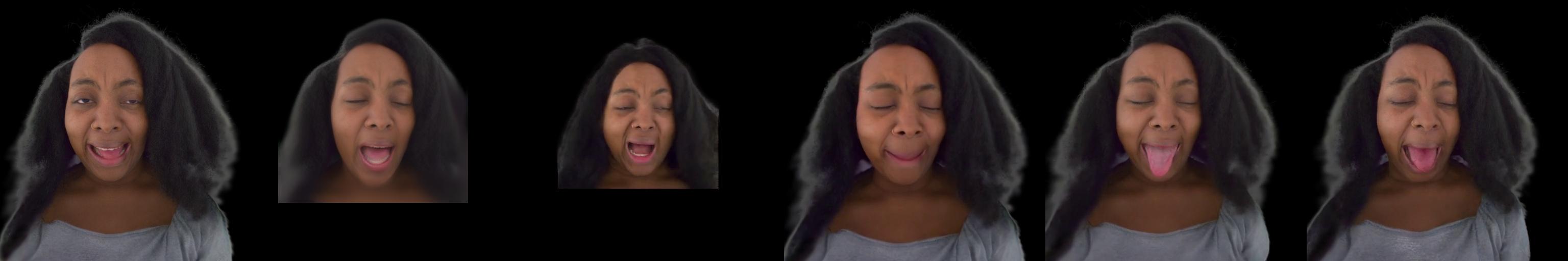}
    \includegraphics[width=\linewidth,trim={0 1.5cm 0 1.5cm},clip]{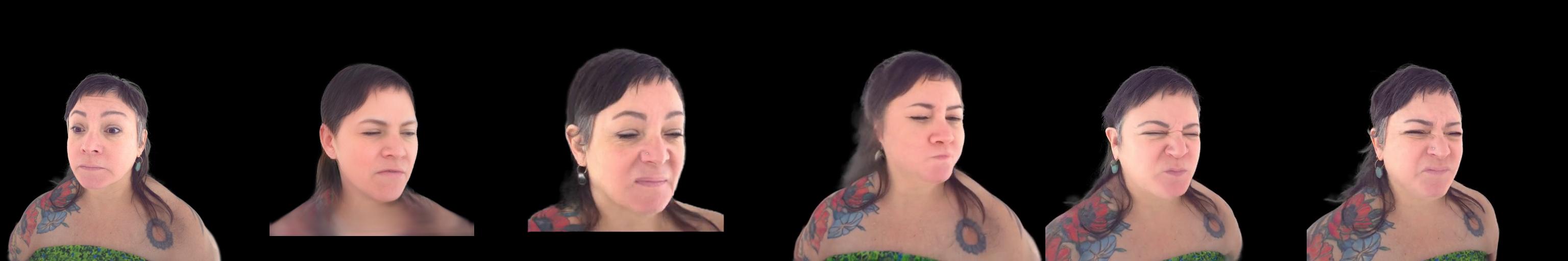}
    \begin{tabular}{p{80pt}p{80pt}p{70pt}p{80pt}p{70pt}p{90pt}}
        \centering Input & \centering  GAGAvatar~\cite{chu2024generalizable}  & \centering CAP4D~\cite{Taubner2024CAP4DCA}  & \centering HunyuanPortrait~\cite{xu2025hunyuanportrait} & \centering \ Ours 
        & \centering GT
    \end{tabular}   
    \vspace{-5mm}
    \caption{\textbf{Comparison against state-of-the-arts on the DynamicSweep Dataset}. The desired output is a video with both view and expression changes. Our method better preserves facial identity, achieves accurate expression control—including challenging cases such as "sticking out tongue"—and provides precise viewpoint control.}
    \label{fig:dynamicsweep_supple}
    \vspace{-5mm}
\end{figure*}

We provide additional qualitative comparisons on Phone Capture, Studio Capture, ViewSweep, and DynamicSweep datasets in Fig.~\ref{fig:mgr_supple}, Fig.~\ref{fig:col_supple}, Fig.~\ref{fig:viewsweep_supple}, and Fig.~\ref{fig:dynamicsweep_supple}, respectively.

\subsection{Cross-Reenactment}
\begin{figure*}
    \centering
    \vspace{-5mm}
   \includegraphics[width=\linewidth,trim={0 2.5cm 0 2.5cm},clip]{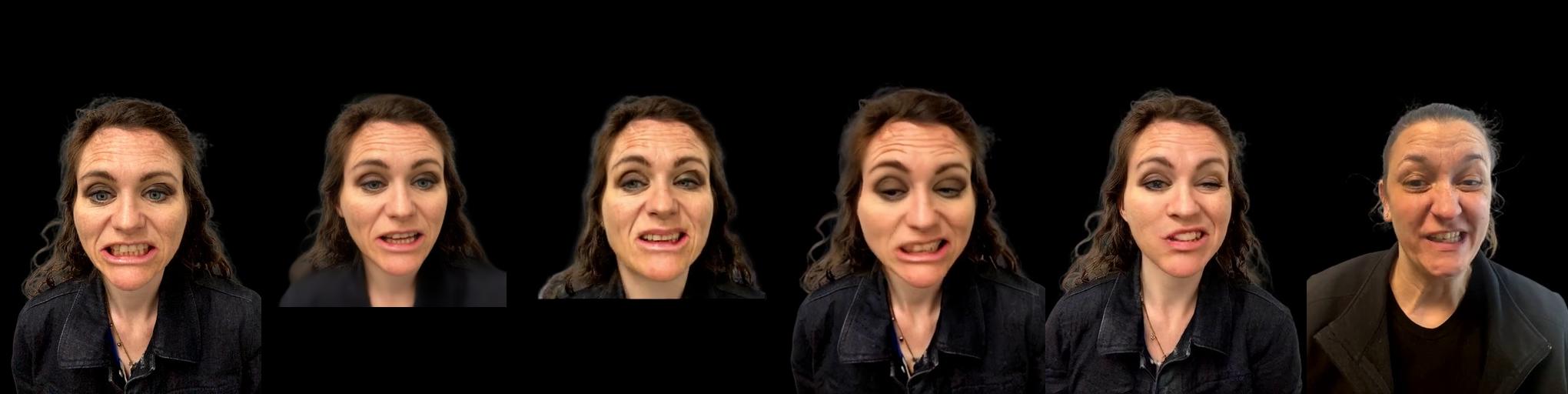}
    \includegraphics[width=\linewidth,trim={0 2.5cm 0 2.5cm},clip]{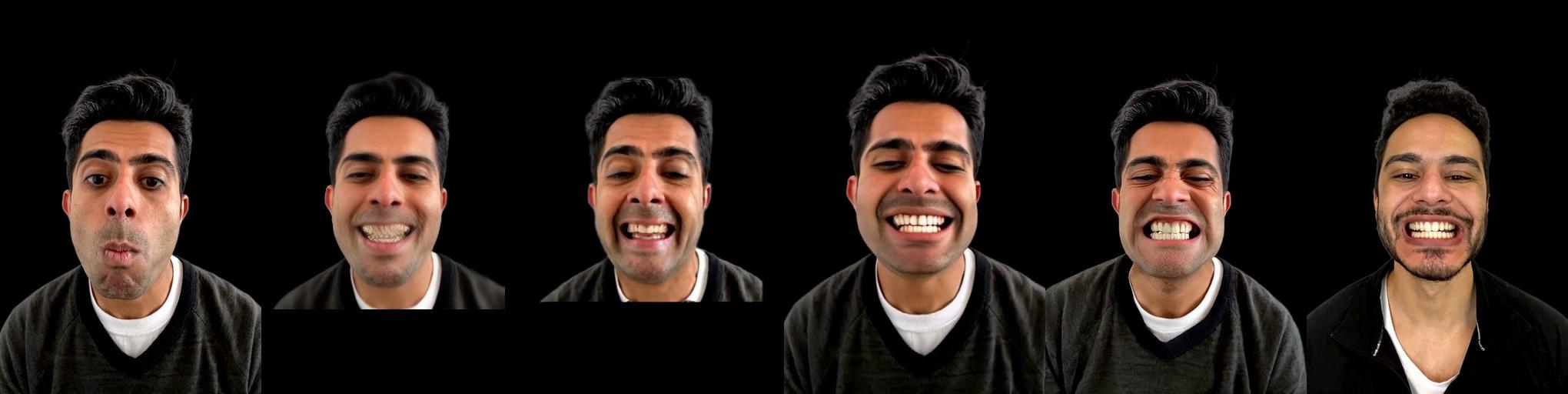}
    \includegraphics[width=\linewidth,trim={0 2.5cm 0 2.5cm},clip]{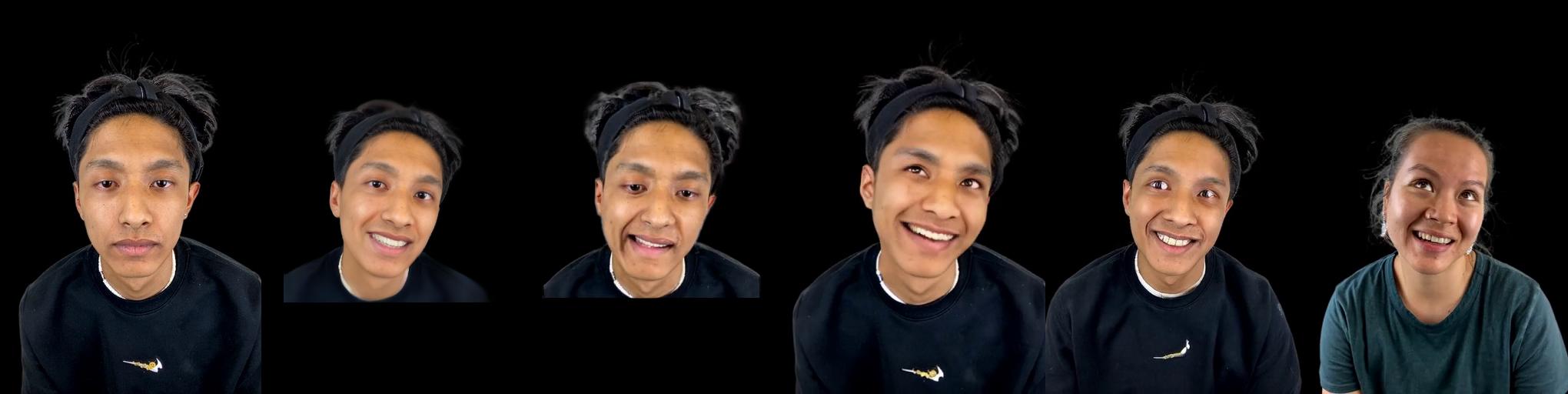}
    \begin{tabular}{p{80pt}p{80pt}p{70pt}p{80pt}p{70pt}p{90pt}}
        \centering Input & \centering  GAGAvatar~\cite{chu2024generalizable}  & \centering CAP4D~\cite{Taubner2024CAP4DCA}  & \centering HunyuanPortrait~\cite{xu2025hunyuanportrait} & \centering \ Ours 
        & \centering Driving
    \end{tabular}   
    \vspace{-5mm}
    \caption{\textbf{Comparison against state-of-the-arts on the task of cross reenactment  on the Phone Dataset}. In this task, we transfer the pose and expressions from the driving identity to the source ID image (input), while enabling continuous viewpoint control. Our method can better preserving facial appearance while enabling precise control over pose and expressions. }
    \label{fig:cross_mgr_supple}
    \vspace{-5mm}
\end{figure*}

\begin{figure*}
    \centering
    \includegraphics[width=\linewidth,trim={0 2.5cm 0 2.5cm},clip]{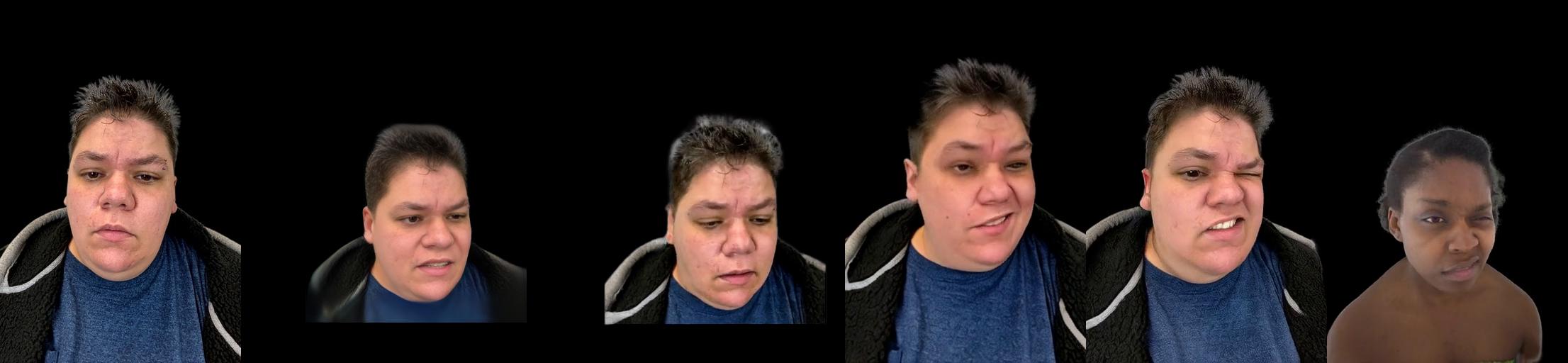}
    \includegraphics[width=\linewidth,trim={0 2.5cm 0 2.5cm},clip]{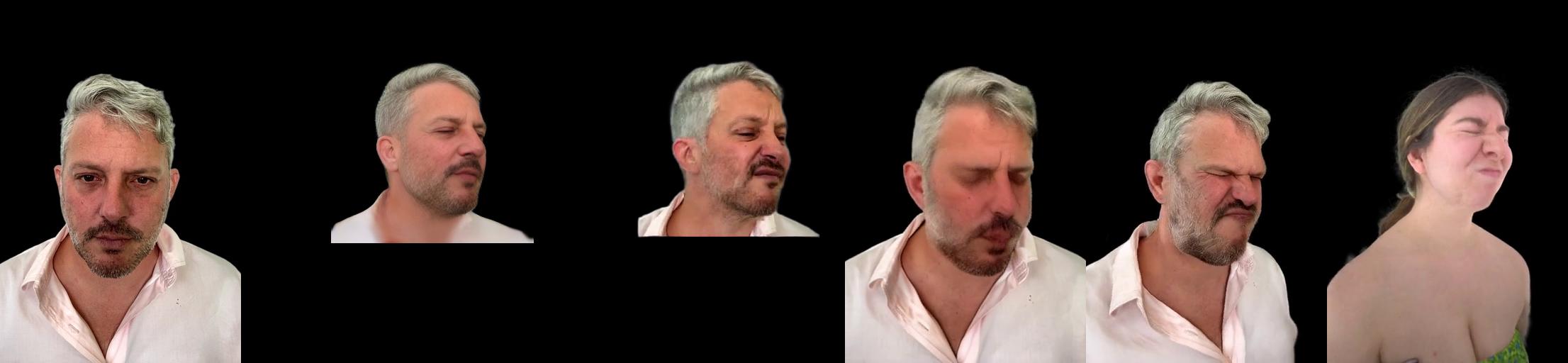}
    \includegraphics[width=\linewidth,trim={0 2.5cm 0 2.5cm},clip]{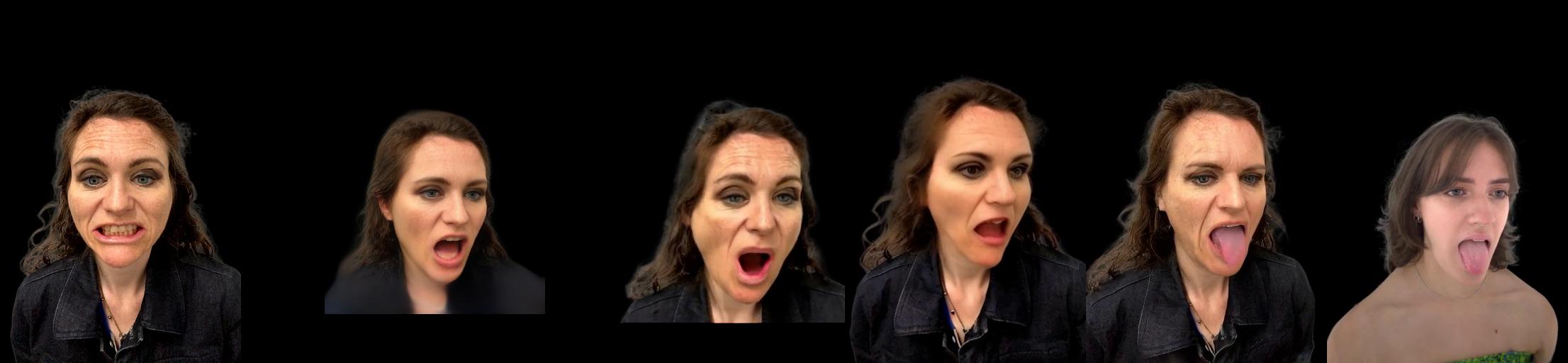}
    \begin{tabular}{p{70pt}p{85pt}p{85pt}p{85pt}p{70pt}p{70pt}}
        \centering Input & \centering  GAGAvatar~\cite{chu2024generalizable}  & \centering CAP4D~\cite{Taubner2024CAP4DCA}  & \centering HunyuanPortrait~\cite{xu2025hunyuanportrait} & \centering \ Ours 
        & \centering GT
    \end{tabular}
    \vspace{-5mm}
    \caption{\textbf{Comparison  against state-of-the-arts on the task of cross reenactment on the DynamicSweep Dataset.} In this task, we transfer the pose and expressions from the driving identity to the source ID image (input), while enabling continuous viewpoint control. Our method achieves more accurate expression control and effectively preserves appearance details from the source ID image. Additionally, it enables precise viewpoint control, resulting in desired novel view synthesis. }
    \label{fig:cross_dynamicsweep_supple}
    \vspace{-5mm}
\end{figure*}

We compare our method against state-of-the-art methods for cross-identity reenactment on Phone Capture dataset (static frontal view, dynamic pose/expression) and Dynamic Sweep dataset (dynamic view, pose and expression). 
The qualitative results are illustrated in Fig.~\ref{fig:cross_mgr_supple} and ~\ref{fig:cross_dynamicsweep_supple}.


\section{Limitations and Future Work}
\label{SecLimit}
While we demonstrate promising video generation results via disentangled expression, pose, and camera control, several limitations remain. First, our method focuses on upper-body portrait generation and does not model hand or full-body animation. Second, similar to other DiT-based video diffusion models, our method faces computational bottlenecks that prevent real-time inference, restricting its use in interactive applications. Third, our current method does not disentangle lighting conditions, which would enable explicit control over illumination and further enhance relighting. We leave these directions for future work.


\end{document}